\pdfoutput=1

\documentclass[journal]{IEEEtran}
\usepackage{cite}

\ifCLASSINFOpdf
   \usepackage[pdftex]{graphicx}
\else
   \usepackage[dvips]{graphicx}
\fi
\usepackage{amsmath}
\usepackage{array}

\ifCLASSOPTIONcompsoc
  \usepackage[caption=false,font=normalsize,labelfont=sf,textfont=sf]{subfig}
\else
  \usepackage[caption=false,font=footnotesize]{subfig}
\fi
\usepackage{dblfloatfix}

\ifCLASSOPTIONcaptionsoff
  \usepackage[nomarkers]{endfloat}
 \let\MYoriglatexcaption\caption
 \renewcommand{\caption}[2][\relax]{\MYoriglatexcaption[#2]{#2}}
\fi
\usepackage{url}

\usepackage{color}
\usepackage{amsmath}
\usepackage{subfig}

\definecolor{orange}{RGB}{255, 192, 0} 

\begin{document}
\onecolumn
\copyright 20XX IEEE.  Personal use of this material is permitted.  Permission from IEEE must be obtained for all other uses, in any current or future media, including reprinting/republishing this material for advertising or promotional purposes, creating new collective works, for resale or redistribution to servers or lists, or reuse of any copyrighted component of this work in other works.
\twocolumn

%
\title{Stock Prices Prediction using Deep Learning Models}
%
%
%

\author{Jialin~Liu,
        Fei~Chao,~\IEEEmembership{Member,~IEEE,}
        Yu-Chen Lin,
        and~Chih-Min~Lin,~\IEEEmembership{Fellow,~IEEE,} 
\thanks{J. Liu and  F. Chao are with the Cognitive Science Department, School of Information Science and Engineering, Xiamen University, China e-mail: (fchao@xmu.edu.cn). Y.-C. Lin is with Department of Accounting, National Chung Hsing University, Taiwan, R.O.C e-mail: (yuchenlin08@gmail.com). C.-M. Lin is with the Department of Electrical Engineering and Innovation Center for Biomedical and Healthcare Technology, Yuan Ze University, Chung-Li, Tao-Yuan 320, Taiwan, R.O.C e-mail: (cml@saturn.yzu.edu.tw). Corresponding Author: Chih-Min Lin}
\thanks{Manuscript received April 19, 2005; revised August 26, 2015.}}

%
%

\markboth{Journal of \LaTeX\ Class Files,~Vol.~14, No.~8, August~2015}%
{Shell \MakeLowercase{\textit{et al.}}: Bare Demo of IEEEtran.cls for IEEE Journals}
%



\maketitle

\begin{abstract}
Financial markets have a vital role in the development of modern society. They allow the deployment of economic resources. Changes in stock prices reflect changes in the market. In this study, we focus on predicting stock prices by deep learning model. This is a challenge task, because there is much noise and uncertainty in information that is related to stock prices. So this work uses sparse autoencoders with one-dimension (1-D) residual convolutional networks which is a deep learning model, to de-noise the data. Long-short term memory (LSTM) is then used to predict the stock price. The prices, indices and macroeconomic variables in past are the features used to predict the next day's price. Experiment results show that 1-D residual convolutional networks can de-noise data and extract deep features better than a model that combines wavelet transforms (WT) and stacked autoencoders (SAEs). In addition, we compare the performances of model with two different forecast targets of stock price: absolute stock price and price rate of change. The results show that predicting stock price through price rate of change is better than predicting absolute prices directly.
\end{abstract}

\begin{IEEEkeywords}
stock, deep learning, LSTM, SAEs
\end{IEEEkeywords}

%
\IEEEpeerreviewmaketitle

\section{Introduction}
%
%
%
%
\IEEEPARstart{S}{tock} time series forecast is one of the main challenges for machine learning technology because the time series analysis is required \cite{tay2001application}. Two methods are usually used to predict financial time series: machine learning models and statistical methods \cite{wang2011forecasting}.

Statistical methods can be used to predict a financial time series. The common methods are autoregressive conditional heteroscedastic (ARCH) methods \cite{engle1982autoregressive}, and autoregressive moving average (ARMA) \cite{box2015time} or an autoregressive integrated moving average (ARIMA) methods. However, traditional statistical methods generally assume that the stock time series pertains to a linear process, and model the generation process for a latent time series to forecast future stock prices \cite{kumar2013performance}. However, a stock time series is generally a dynamic nonlinear process \cite{si2013obst}.

Many machine learning models can capture nonlinear characters in data without prior knowledge \cite{atsalakis2009surveying}. These models are always used to model a financial time series. The most commonly used models for stock forecasts are artificial neural networks (ANN), support vector machines (SVM), and hybrid and ensemble methods. Artificial neural networks have found many applications in business because they can deal with data that is non-linear, non-parametric, discontinuous or chaotic for a stock time series \cite{liu2012fluctuation}. Support vector machine is a statistical machine learning model that is widely applied for pattern recognition. A SVM model, which learns by minimizing the risk function and the empirical error and regularization terms has been derived to minimize the structural risk \cite{chen2010svm}. Box et el. presented a revised least squares (LS)-SVM model and predicted movements in the Nasdaq Index after training with satisfactory results \cite{box2015time}.

Deep learning models, which are an extension of ANN¡¯s, have seen recent rapid development. Many studies use deep learning to predict financial time series. For example, Ting et al. used a deep convolutional neural network to forecast the effect of events on stock price movements \cite{ding2015deep}. Bengio et al. used long-short term memory (LSTM) to predict stock prices \cite{baek2018modaugnet}.

This study addresses the problem of noise in a stock time series. Noise and volatile features in a stock price forecast are major challenges because they hinder the extraction of useful information \cite{wang2012novel}. A stock time series can be considered as waveform data, so the technology from communication electronics such as wavelet transform is pertinent. Bao et al. used a model that combines wavelet transform and stacked autoencoder (SAE) to de-noise a financial time series \cite{bao2017deep}. This study de-noises data using an autoencoder \cite{hinton2006reducing,bengio2007greedy} with a convolutional resident neural network (Resnet) \cite{he2016deep}. This is an adaptive method to reduce noise and dimension for time sequences. It is different from wavelet transforms in that the kernel of the convolutional neural network adapts to dataset automatically, so it can more effectively eliminate noise and retain useful information. The experiments use the CSI 300 index, the Nifty 50 index, the Hang Seng index, the Nikkei 225 index, the S\&P 500 index and the DJIA index are performed and the results are compared with those for \cite{bao2017deep}. The proposed model gives more accurate predictions, as measured by mean absolute percent error (MAPE), Theil U and the linear correlation between the predicted prices and the real prices. We do both the experiments on predicting stock price directly and on predicting price rate of change and calculating the price indirectly. We found that the latter can achieve better accuracy. Predicting future price indirectly can be seen as adding prior knowledge to improve model performance.

The remainder of this paper has five sections. The next section draws the background knowledge of market analysis. Section \ref{De-noising_CNN} details a little experiment about the property of de-noising CNN. Section \ref{Methodology} details the structure for the proposed model with sparse autoencoders and LSTM. Section \ref{Experiment} describes the features and data resources for the experiment and details the experiment, and analyzes the results of the experiment. The last section draws conclusions.

\section{Background}\label{Background}
Understanding the behaviors of the market in order to improve the decisions of investors is the main purpose of market analysis. Several market attributes and features that are related to stock prices time series have been studied. Depending on the market factors that are used, market analysis can be divided into two categories: fundamental and technical analysis \cite{cavalcante2016computational}.

Technical analysis often only uses historical prices as market characters to identity the pattern of price movement. Studies assume that the relative factors are incorporated in the movement of the market price and that history will repeat itself. Some investors used technical approaches to predict stock prices with great success \cite{rodriguez2011cast}. However, the Efficient Market Hypothesis \cite{fama1965behavior} assumes that all available factors are already incorporated in the prices so only new information affects the movement of market prices, but new information is unpredictable.

Fundamental analysis assumes that the related factors are the internal and external attributes of a company. These attributes include the interest rate, product innovation, the number of employees, the management policy and etc \cite{lam2004neural}. In order to improve the prediction, other information such as the exchange rate, public policy, the Web and financial news are used as features. Nassirtoussi et al. used news headlines as features to predict the market \cite{nassirtoussi2015text}. Twitter sentiment was used in \cite{porshnev2013machine} to improve predictions.

In 1995, one study showed that 85\% of responders depend on fundamental analysis and technical analysis \cite{lui1998use}. Technical analysis is more useful for short-term forecasting so it is pertinent to high frequency trading. Lui et al. showed that technical analysis better forecasts turning points than trends, but fundamental analysis gives a better prediction of trends \cite{lui1998use}.

Depending on the prediction target, tasks can be classified as regression task or classification tasks. For a regression task the prediction target for the model is the future price, and a classification task model predicts the rise or fall of the stock prices. If the predicted price is higher than the current price, the recommended strategy is to buy, and vice versa. This is the buy-and-sell trading strategy, which is widely used in studies \cite{yao1999neural}. If the task is to identify the rise or fall in the price, then the resultant strategy is obvious. Market analysis is also used for recommendation systems. Huang et al. used SVR to predict the return of each stock and to select stocks with the highest profit margins (top 10, 20 and 30) to calculate the profit margin \cite{huang2012hybrid}.

This study uses technical analysis to predict the stock price for the next day. Sparse autoencoders with 1-D convolution networks and prior knowledge are used to give a more accurate prediction than other techniques.

\section{De-noising CNN}\label{De-noising_CNN}
To create a 1-D convolutional neural network for sequence analysis, a single neural network can be combined with a convolutional neural network with LSTM. When the features for the input are extracted at a high-level by the convolution layer, the price is directly predicted by the LSTM layer. During training, the gradient propagates back to convolution through the LSTM layer. However, if there is too much noise in the data, this model tends to over-fit.

\begin{figure}
\begin{center}
\subfloat[Training and test loss]{
\includegraphics[width=78.5mm]{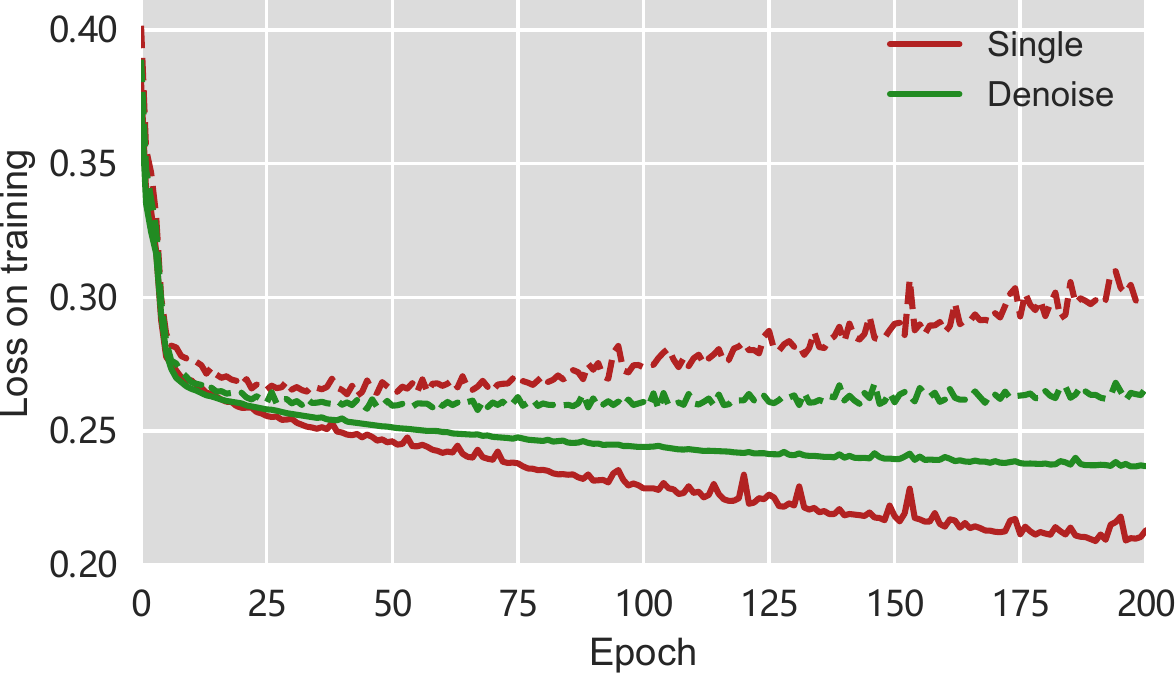}
\label{CNN_overfit1}}

\subfloat[Loss gap]{
\includegraphics[width=78.5mm]{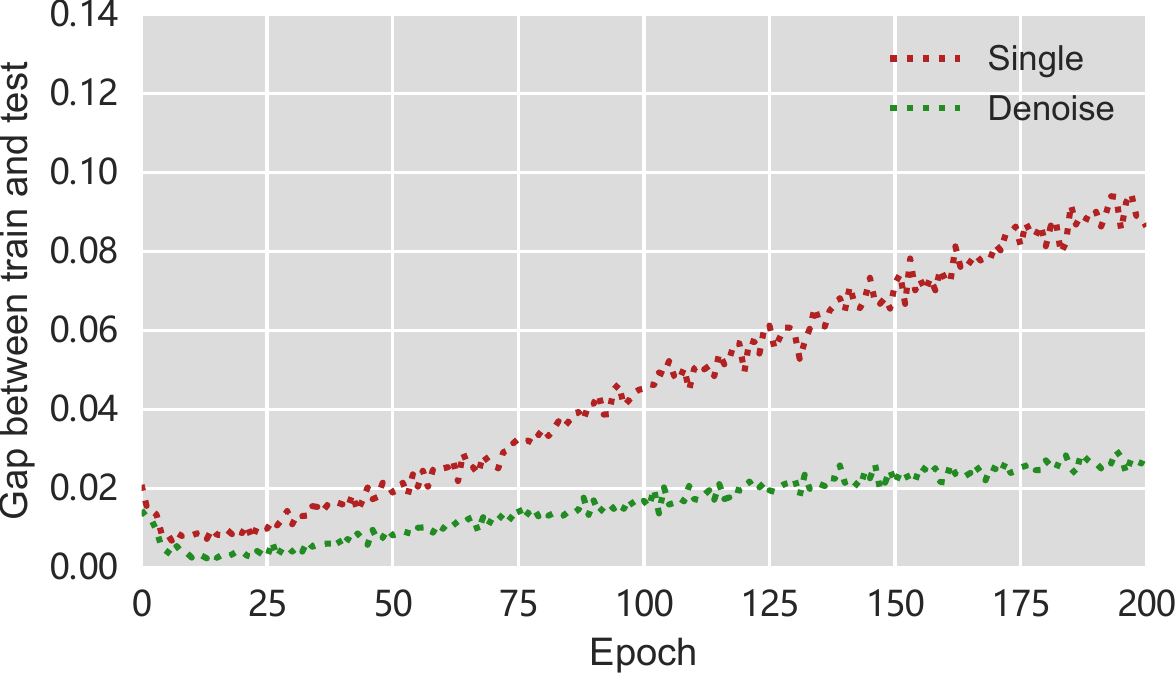}
\label{CNN_overfit2}}
\end{center}
   \caption{Training curve.}
\label{CNN_overfit}
\end{figure}

A notional problem is used to compare the model with a single neural network. The model uses the features after de-noising. The task is a bias prediction task, in which each data point corresponds to a function, $\mathrm{y}=\sin(\mathrm{x}+2\pi*b)$. The target is to predict the value of $b$ in this function, which is sampled from a uniform distribution, $U(-1,1)$. Here $\mathrm{y}$ is the feature vector for the data, where $\mathrm{x}=[-2\pi,-2\pi+\frac{4}{m}\pi,\cdots,-2\pi+\frac{4(m-1)}{m}\pi]^\mathrm{T}$, $m$ is the size of sequence. Two types of noises are then added to the features. The first type is Gaussian noise, $n_G\sim\mathcal{N}(\mu,\delta^2)$. The form of another noise is written as $\lambda\sum_i^nc_i\exp(x-b_{ri})^2$, where $b_{ri}$ is sampled from the uniform distribution, $U(-1,1)$, $c_i$ is sampled from the 0-1 distribution $B(1,0.5)$ with possibility $p=0.5$ and $\lambda$ is the weight of this noise. This noise has multiple peaks that interfere with prediction. Figure \ref{CNN_overfit1} shows the training curves for both models. The red and green lines are the respective training curves for the model that combines CNN with LSTM and uses the features after de-noising. The solid and dashed lines respectively represent the training loss and the test loss. In Figure \ref{CNN_overfit2} the dotted curves indicate the loss gap. When the training loss decreases, the loss gap for the model grows slower than that for the single neural network. The minimum test loss for the proposed model is less than that for the single neural network. It is obvious that de-noising features prevents over-fitting for the model.

\begin{figure}
\begin{center}
\includegraphics[width=78.5mm]{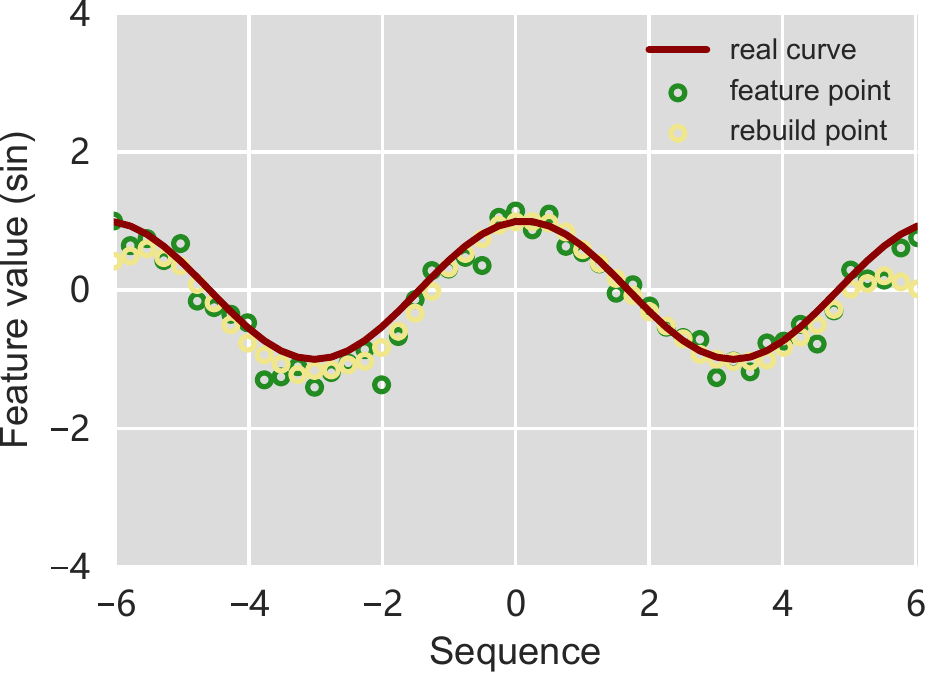}
\end{center}
   \caption{Sine curve rebuilding.}
\label{CNN_rebuild}
\end{figure}

The noise for stock forecasting is much more complex than the noise for this notional task, so in this study the noise in the stock forecast data is reduced first using 1-D convolution autoencoders. The details of the features of the 1-D CNN autoencoders processes are given. In Figure \ref{CNN_rebuild}, the yellow dots denote the rebuilt curve for the sine function. The red curve is the global true, which is the sine function curve without noise, and the green dots are feature points with noise. The ordinate axis represents the specific feature value. Each point represents an element input for the model. It is obvious that curve for the yellow dots is smoother than that for the green dots and it is close to the real curve.

\begin{figure}
\begin{center}
\subfloat[Kernel in Single]{\includegraphics[width=65mm]{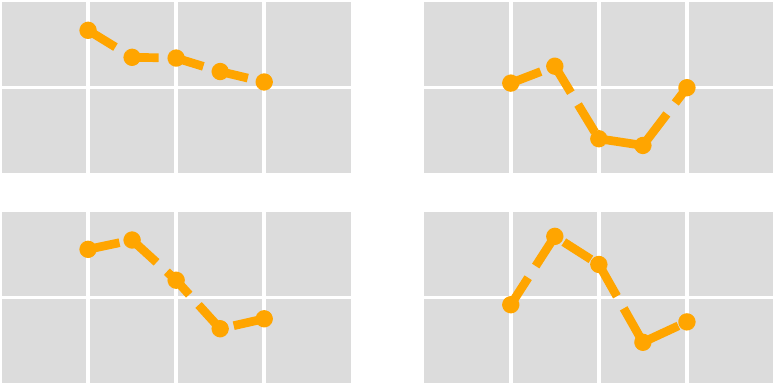}}

\subfloat[Kernel in Autoencoders]{\includegraphics[width=65mm]{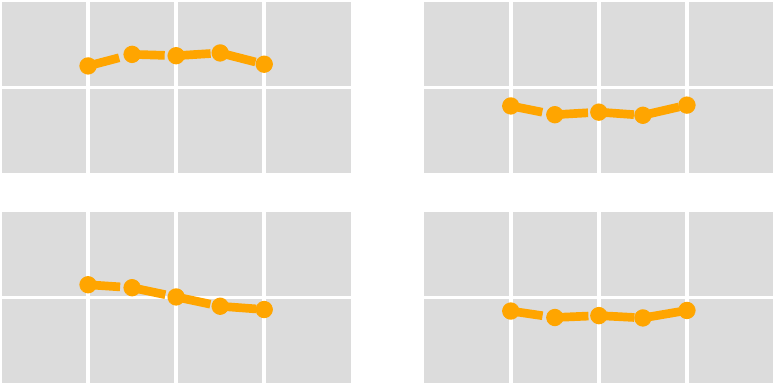}}
\end{center}
   \caption{Convolution kernel of model. }
\label{CNN_kernel}
\end{figure}

The values for the weights in the convolutional kernel are shown in Figure \ref{CNN_kernel}, which is for the model with minimal test loss. The values for the weights in the convolutional kernel are also smoother than those for a single neural network (see Figure \ref{CNN_kernel}). However, the sine function is smoother than the noise, so the kernel in the single network is more likely to match the noise than the 1-D convolution autoencoders. This model tries to establish a relationship between the noise and the label. In fact, the noise and the label are irrelevant, so it is more prone to over-fitting.

\section{Methodology}\label{Methodology}
In order to extract high-level abstract features and predict future prices from the stock time series, we apply two models in our system, one deep model is used for de-noising and another is used for prediction. The prediction process involves three steps:(1) data preprocessing that involves calculating technical indicators, clipping and normalizing features, (2) encoding and decoding features using a 1-D ResNet block to minimize the rebuilt loss and (3) using the LSTM to deal with high-level abstract features and give a one-step-ahead output.

\begin{figure*}
\begin{center}
\includegraphics[width=180.9mm,height=189.6mm]{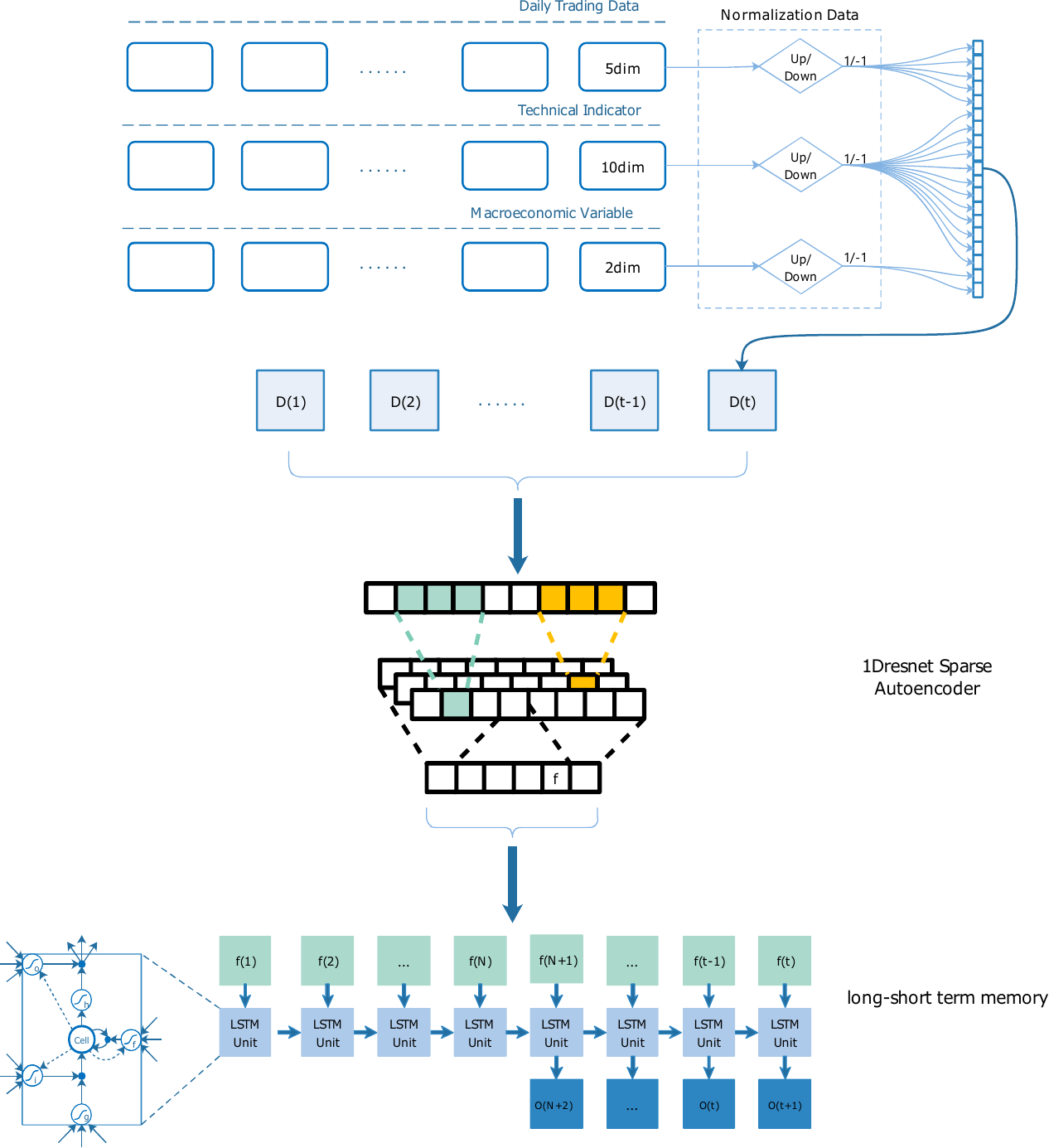}
\end{center}
   \caption{Three steps process of high-level abstract features extraction and prediction.}
\label{StockPredictionFlow}
\end{figure*}

Figure \ref{StockPredictionFlow} shows the overall framework. The input feature of data sequence is a $c\times t$ matrix, where $c$ is the number of channels, and $t$ is the length of sequence. Daily trading data, technical indicators and macroeconomic variables are the matrices of data sequence with size $5\times t$, $10\times t$ and $2\times t$. After preprocessing, we merge them into one matrix with size $17\times t$, so the inputted data sequence has 17 channels. The prices are then predicted by LSTM after the noise and dimension have been reduced by the encoder model.

\subsection{Sparse autoencoders}
\begin{figure}[ht!]
\begin{center}
\includegraphics[width=68.81mm,height=76mm]{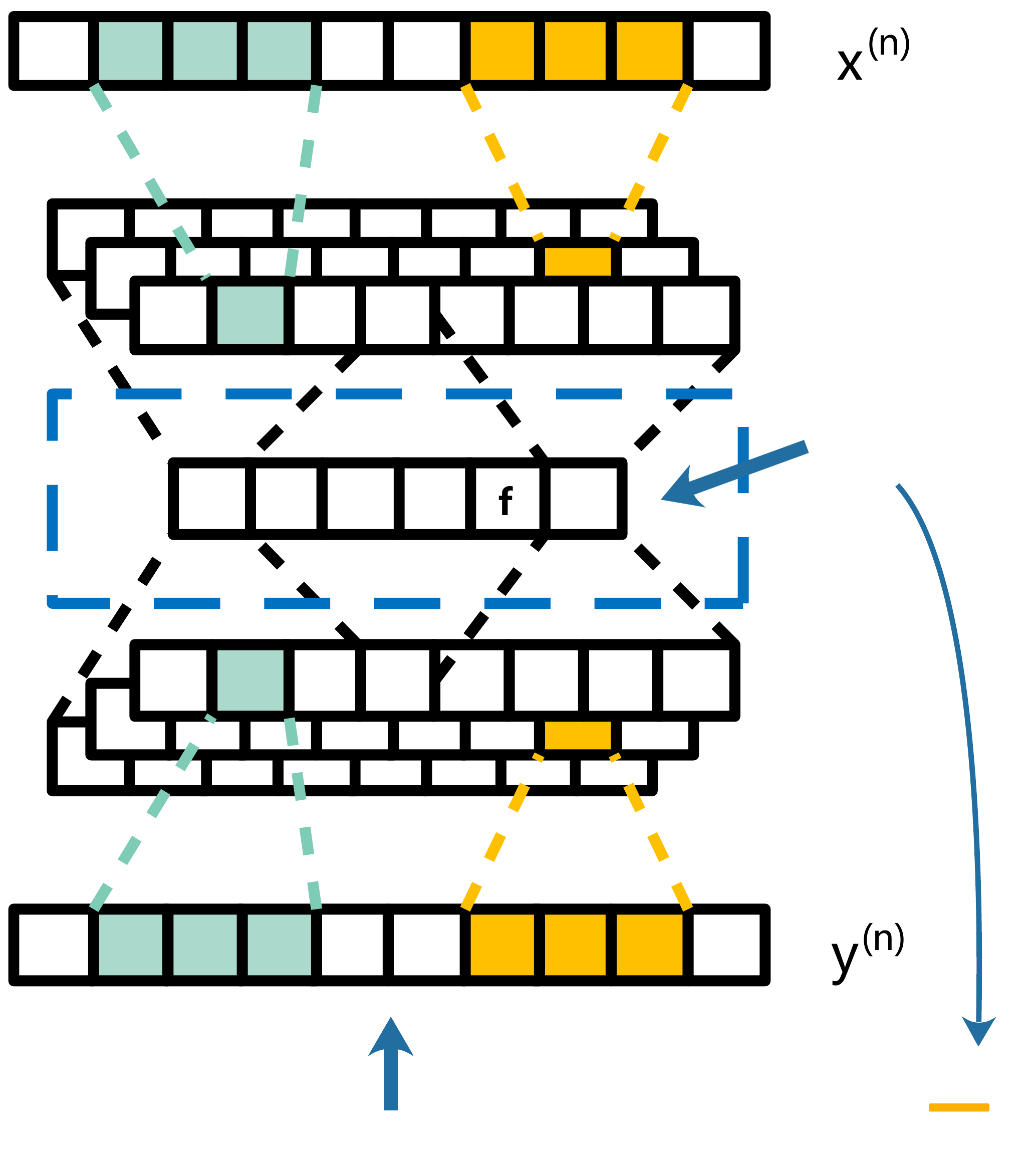}
\put(-46,130){$\sum\limits^S_jKL(\rho\|\hat{\rho}_j)$}
\put(-120,12){\small $\frac{1}{N}\sum\limits^N_n\frac{1}{2}\|{\rm x}^{(n)}-{\rm y}^{(n)}\|^2+\beta\mathcal{L}_{sp}$}
\end{center}
\caption{Sparse autoencoders with 1 dimension convolution neural network.}
\label{SAEs}
\end{figure}

Sparse autoencoders are models that can reduce the dimension. An autoencoder neural network is used to rebuild the input (see Figure \ref{SAEs}). The loss function, which is used to train autoencoder neural network, is given by \cite{hinton2006reducing,bengio2007greedy}

\begin{equation}
\begin{aligned}
\mathcal{L}=\frac{1}{N}\sum^N_n\frac{1}{2}\|\mathrm{x}^{(n)}-\mathrm{y}^{(n)}\|^2+\beta\mathcal{L}_{sp}
\end{aligned}
\end{equation}
where $N$ is the number of data points, $\mathrm{x}^{(n)}$ denotes the feature vector for the $nth$ sample and $\mathrm{y}^{(n)}$ denotes the reconstructed feature vector for the $nth$ sample. The last term is the sparse penalty term and $\beta_{sp}$ is the weight. The sparse penalty, which is a kind of regularization, is used to make most units of network tend to non-activity state in order to reduce over-fitting. This is the difference between sparse autoencoders and traditional autoencoders. The sparse penalty is given by \cite{andrew20sparse},

\begin{equation}
\begin{aligned}
\mathcal{L}_{sp}&=\sum^S_jKL(\rho\|\hat{\rho}_j)\\
&=\sum^S_j\left[\rho\log\frac{\rho}{\hat{\rho}_j}+(1-\rho)\log\frac{1-\rho}{1-\hat{\rho}_j}\right]
\end{aligned}
\end{equation}
where $\rho$ is the sparse parameter, $S$ is the number of units in the hidden layer and $\hat{\rho}_j=\sigma(x_j)$, $x_j$ is the $jth$ unit in the hidden layer, $\sigma(x)=\frac{1}{1+e^{-x}}$. Weight decay is also used to reduce out-fitting of the model. After training, only the features from the middle layer of the network are used (see Figure \ref{SAEs}).

The model for the sparse autoencoders \cite{hinton2006reducing,bengio2007greedy} is a 1-D CNN. This is used to compare the performance of WT and CNN in terms of de-noising stock time series data. A convolution network is used as the encoding network, and a deconvolution network is the decoded network \cite{zeiler2010deconvolutional}, so the model used in SAEs is a fully convolutional network. The autoencoder's function is not only to reduce noise, but also to reduce the dimensions of the features, in order to allow the latter network structure to use a smaller number of weights. The CNN applied here is the ResNet \cite{he2016deep}, which is a type of convolutional neural network used to speed up the training by using a ``shortcut connections'' \cite{he2016deep} to back-propagate gradient.

\subsection{Long-short term memory}
\begin{figure}
\begin{center}
\includegraphics[width=82.37mm]{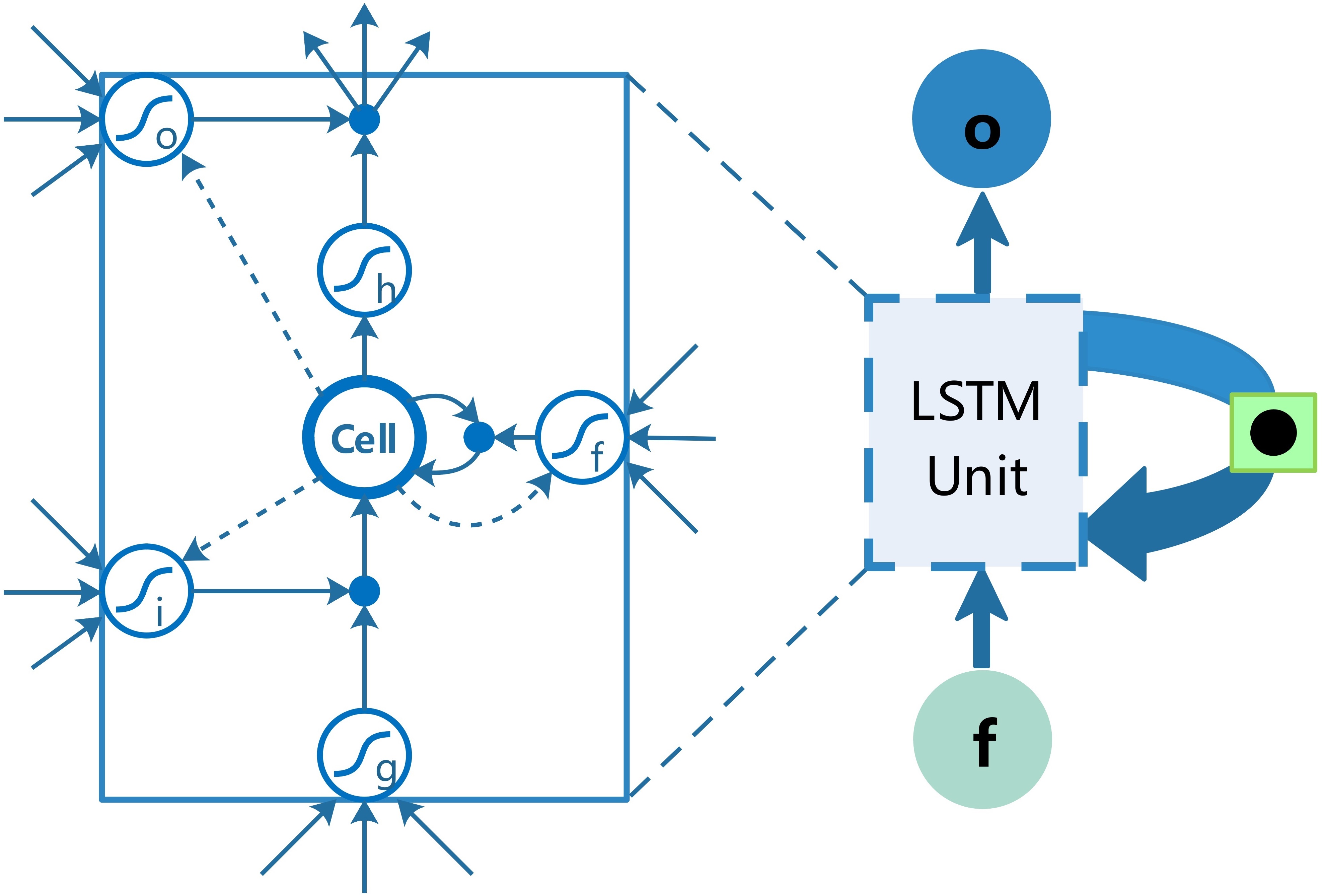}
\end{center}
   \caption{Long-short term memory unit.}
\label{LSTMunit}
\end{figure}

LSTM is a type of recurrent neural network (RNN) \cite{how2016behavior} that can be used to transfer information from the past to the present. However, the structure of a RNN has a defect that can cause the gradient to vanish or explode when the input series are too long. The problem of the gradient exploding is generally solved by gradient clipping

\begin{equation}
\hat{g}=\left\{
\begin{aligned}
&\frac{\hat{g}*threshold}{\|\hat{g}\|},\ if\ \|\hat{g}\|>threshold;\\
&\hat{g}\ \ \ \ \ \ \ \ \ \ \ \ \ \ \ \ \ \ \ \ \ \ ,\ other,
\end{aligned}
\right.
\end{equation}
where $\hat{g}$ represents the gradient of a parameter. The problem of the gradient vanishing is solved by using the structure of the LSTM. A LSTM differs from a conventional RNN in that the LSTM has another memory that transfers its state to the next state without matrix multiplication and operation of activation function, so the gradient is back-propagated smoothly \cite{hochreiter1997long}. The details of the LSTM are shown in Figure \ref{LSTMunit}. The left part of figure is the structure of the LSTM unit. The dotted arrows in the figure indicate the indirect effects. At each step, all the $g,i,f$ and $o$ gates receive the last state and the new feature, and then the cell state and the hidden state are updated at time t, and the input for the unit is the last state vector for the cell ($c_{t-1}$), the hidden last state vector ($h_{t-1}$) and the input feature ($x_t$). The four vectors are
\begin{equation}
\begin{aligned}
g_t=\tanh(\mathrm{W_g}[x_t,h_{t-1}]+b_g)\\
\end{aligned}
\end{equation}
\begin{equation}
\begin{aligned}
i_t=\sigma(\mathrm{W_i}[x_t,h_{t-1}]+b_i)
\end{aligned}
\end{equation}
\begin{equation}
\begin{aligned}
f_t=\sigma(\mathrm{W_f}[x_t,h_{t-1}]+b_f)
\end{aligned}
\end{equation}
\begin{equation}
\begin{aligned}
o_t=\sigma(\mathrm{W_o}[x_t,h_{t-1}]+b_o)
\end{aligned}
\end{equation}
where $\sigma(x)=\frac{1}{1+e^{-x}}$, and $g_t$ is the new information that is used to update the cell state, and $i_t$ and $f_t$ are respectively used to select information that is to be added to cell state or be forgotten,

\begin{equation}
\begin{aligned}
c_t=i_t*g_t+f_t*c_{t-1}
\end{aligned}
\end{equation}
where $*$ denotes element-wise multiplication. The term $o_t$ is used to select the output and the hidden state,
\begin{equation}
\begin{aligned}
h_t=o_t*\tanh(c_t)
\end{aligned}
\end{equation}
then $output_t=h_t$.

\section{Experiment}\label{Experiment}
The experiments compare the accuracy of the proposed method with that of a deep learning framework \cite{bao2017deep} for the CSI 300 index, the DJIA index, the Hang Seng index, the Nifty 50 index, the Nikkei 225 index and the S\&P500 index. Similar to a previous study \cite{bao2017deep}, more than one market is used. The predictive accuracy is evaluated by MAPE, Theil U and the linear correlation between the prediction and the real price \cite{guo2014feature,hsieh2011forecasting,altay2005stock,emenike2010forecasting}. The data is divided into different groups for training and testing, in order to reduce the time span.

Two experiments test the performance of the two methods: (1) a 1-D resnet autoencoder is used to predict prices (called C1D-LSTM) and (2) a 1-D resnet autoencoder is used to predict the rate of change of prices (called C1D-ROC). The accuracy of the models is compared and the prediction curve for one year is plotted.

\subsection{Data descriptions}
\textbf{Data resource.} The data resource is following a previous study \cite{bao2017deep} from the Figshare website. The data was sampled from the WIND(\url{http://www.wind.com.cn}) and CSMAR(\url{http://www.gtarsc.com}) databases of the Shanghai Wind Information Co., Ltd and the Shenzhen GTA Education Tech. Ltd, respectively. The stock time series is from $1^{\rm st}$ Jul. 2008 to $30^{\rm th}$ Sep. 2016 (see Table \ref{Time_Interval}).

\textbf{Data features.} Following a previous study \cite{bao2017deep}, three sets of features are selected as the inputs. The first set is the trading data for the past, including Opening, Closing, High, and Low prices and trading volume. In Table \ref{Technical_Indicator}, $C_t$, $L_t$ and $H_t$ respectively denote the closing price, the low price and the high price at time t. The second set includes the technical indicators that are widely used for stock analysis. Their calculation method is shown in Table \ref{Technical_Indicator}, where $DIFF_t=EMA(12)_t-EMA(26)_t$, $Ds$ and $Dhl$ respectively denote the double exponential moving average for $C-\frac{HH+LL}{2}$ and $HH-LL$, where $HH$ and $LL$ respectively denote the highest high price and the lowest low price in the range. The last set of features is the macroeconomic information. Stock prices are affected by many factors, so using the macroeconomic information as features can reduce uncertainty in the stock prediction. The US dollar index and the Interbank offered rate for each market are the third set of features.

\textbf{Data divide.} The data is divided to train multiple models. Each model is trained using past data, and the training data and test data cannot be randomly sampled from the dataset because it is irrational. To predict future stock prices, only data from the past can be used. The greater the time interval between the two stock time series data, the smaller is the correlation between them; so using outdated data does not improve performance. In order to take into account the above reason and to simplify the result, the forecast is divided into 6 years; and each year is from $1^{\rm st}$ Oct. to $30^{\rm th}$ Sep. (see Table \ref{Time_Interval}).

\renewcommand\arraystretch{1.}
\begin{table}
\caption{The prediction time interval of each year.}
\begin{center}
\begin{tabular}{c|l}
\hline
Year & \multicolumn{1}{c}{Time Interval} \\
\hline
1th & 2010.10.01$\sim$2011.09.30 \\
2th & 2011.10.01$\sim$2012.09.30 \\
3th & 2012.10.01$\sim$2013.09.30 \\
4th & 2013.10.01$\sim$2014.09.30 \\
5th & 2014.10.01$\sim$2015.09.30 \\
6th & 2015.10.01$\sim$2016.09.30 \\
\hline
\end{tabular}
\end{center}
\label{Time_Interval}
\end{table}

\renewcommand\arraystretch{1.5}
\begin{table*}[ht!]
\caption{The Technical Indicator used in experiment following \cite{bao2017deep}.}
\begin{center}
\begin{tabular}{c|l|l}
\hline
Name& \multicolumn{1}{c}{Definition} & \multicolumn{1}{|c}{Formulas} \\
\hline
MACD & Moving Average Convergence & $MACD(n)_{t-1}+\frac{2}{n+1}\times(DIFF_t-MACD(n)_{t-1})$\\
CCI & Commodity channel index & $\frac{M_t-SM_t}{0.015D_t}$\\
ATR & Average true range & $\frac{1}{n}\sum^n_{i=1}TR_i$\\
BOLL & Bollinger Band MID & ${\rm MA}20$\\
EMA20 & 20 day Exponential Moving Average &$\frac{2}{21}\times(C_t-EMA_{t-1})+(1-\frac{2}{21})\times EMA_{t-1}$\\
MA5/MA10 & 5/10 day Moving Average & $\frac{C_t+C_{t-1}+\cdots+C_{t-4}}{5}/\frac{C_t+C_{t-1}+\cdots+C_{t-9}}{10}$\\
MTM6/MTM12 & 6/12 month Momentum &$C_t-C_{t-6}/C_t-C_{t-12}$\\
ROC & Price rate of change &$\frac{C_t-C_{t-N}}{C_{t-N}}*100$\\
SMI & Stochastic Momentum Index & $\frac{Ds}{Dhl}*100$\\
WVAD & Williams's Variable Accumulation/Distribution & $AD_{t-1}+\frac{(C_t-L_t)-(H_t-C_t)}{H_t-C_t}*volume$\\
\hline
\end{tabular}
\end{center}
\label{Technical_Indicator}
\end{table*}

\subsection{Evaluation}
The experiments use MAPE,the linear correlation between the predicted price and the real price and Theil U to evaluate the model. These are defined as

\begin{equation}
\begin{aligned}
\mathrm{MAPE}=\frac{1}{N}\sum^N_{t=1}\left|\frac{y_t-y_t^*}{y_t}\right|
\end{aligned}
\end{equation}

\begin{equation}
\begin{aligned}
\mathrm{R}=\frac{\sum^N_{t=1}(y_t-\overline{y_t})(y^*_t-\overline{y^*_t})}{\sqrt{\sum^N_{t=1}(y_t-\overline{y_t})^2\sum^N_{t=1}(y^*_t-\overline{y^*_t})^2}}
\end{aligned}
\end{equation}

\begin{equation}
\begin{aligned}
\mathrm{Theil\ U}=\frac{\sqrt{\frac{1}{N}\sum^N_{t=1}(y_t-y_t^*)^2}}{\sqrt{\frac{1}{N}\sum^N_{t=1}(y_t)^2}+\sqrt{\frac{1}{N}\sum^N_{t=1}(y^*_t)^2}}
\end{aligned}
\end{equation}
where $y_t$ and $y^*_t$ respectively denote the predictive price for the proposed model and the actual price on day $t$, and $\overline{y_t}$ and $\overline{y^*_t}$ respectively denote their average values. MAPE is a measure of the relative error in the average values. R is the correlation coefficient for two variables and describes the linear correlation between them. A large value for R means that the forecast is close to the actual value. Theil U is also called the uncertainty coefficient and is a type of association measure. A smaller value for MAPE and Theil U denotes greater accuracy.

\subsection{Predictive accuracy test}
Tables \ref{CSI_300}-\ref{SP500} show that a 1-D CNN gives slightly better results than WSAEs. This shows that the convolutional network is effective in processing stock data, which is a model that can adaptively de-noise the noisy data and can reduce the dimensionality. Markets with higher predicted errors are almost the same for both two models. Moreover, the CSI 300 index, the HangSeng Index and the Nifty 50 index are more difficult to be predicted than the DJIA index and the S\&P500 Index.

In some individual cases, more closer between predicted and actual prices does not mean that there is a higher prediction accuracy. However, the average for different years shows that the prediction accuracy and the linear correlation are positively correlated.

\renewcommand\arraystretch{1.}
\begin{table}[ht!]
\caption{The prediction accuracy in CSI 300 index.}
\begin{center}
\resizebox{1.\hsize}{!}{
\begin{tabular}{c|c|c|c|c|c|c|c}
\hline
Year & Year1 & Year2 & Year3 & Year4 & Year5 & Year6 & Average \\
\hline
\multicolumn{8}{c}{Panel A.MAPE} \\
\hline
WSAEs-LSTM & 0.025 & 0.014 & 0.016 & 0.011 & 0.033 & 0.016 & 0.019 \\
C1D-LSTM & 0.015 & 0.014 & 0.017 & 0.011 & 0.051 & 0.015 & 0.020 \\
C1D-ROC & 0.015 & 0.011 & 0.013 & 0.009 & 0.025 & 0.012 & 0.014  \\
\hline
\multicolumn{8}{c}{Panel B.Correlation coefficient} \\
\hline
WSAEs-LSTM & 0.861 & 0.959 & 0.955 & 0.957 & 0.975 & 0.957 & 0.944 \\
C1D-LSTM & 0.961 & 0.960 & 0.951 & 0.961 & 0.976 & 0.959 & 0.961  \\
C1D-ROC & 0.957 & 0.969 & 0.959 & 0.974 & 0.987 & 0.969 & 0.969  \\
\hline
\multicolumn{8}{c}{Panel C.Theil U} \\
\hline
WSAEs-LSTM & 0.017 & 0.009 & 0.011 & 0.007 & 0.023 & 0.011 & 0.013 \\
C1D-LSTM & 0.009 & 0.009 & 0.011 & 0.007 & 0.031 & 0.011 & 0.013  \\
C1D-ROC & 0.010 & 0.007 & 0.010 & 0.006 & 0.017 & 0.009 & 0.010  \\
\hline
\end{tabular}
}
\end{center}
\label{CSI_300}
\end{table}

\begin{table}[ht!]
\caption{The prediction accuracy in DJIA index.}
\begin{center}
\resizebox{1.\hsize}{!}{
\begin{tabular}{c|c|c|c|c|c|c|c}
\hline
Year & Year1 & Year2 & Year3 & Year4 & Year5 & Year6 & Average \\
\hline
\multicolumn{8}{c}{Panel A.MAPE} \\
\hline
WSAEs-LSTM & 0.016 & 0.013 & 0.009 & 0.008 & 0.008 & 0.010 & 0.011 \\
C1D-LSTM & 0.011 & 0.010 & 0.010 & 0.007 & 0.010 & 0.011 & 0.010 \\
C1D-ROC & 0.011 & 0.008 & 0.007 & 0.007 & 0.009 & 0.008 & 0.008 \\
\hline
\multicolumn{8}{c}{Panel B.Correlation coefficient} \\
\hline
WSAEs-LSTM & 0.922 & 0.928 & 0.984 & 0.952 & 0.953 & 0.952 & 0.949 \\
C1D-LSTM & 0.958 & 0.964 & 0.982 & 0.975 & 0.939 & 0.953 & 0.962 \\
C1D-ROC & 0.953 & 0.975 & 0.988 & 0.969 & 0.946 & 0.972 & 0.967  \\
\hline
\multicolumn{8}{c}{Panel C.Theil U} \\
\hline
WSAEs-LSTM & 0.010 & 0.009 & 0.006 & 0.005 & 0.005 & 0.006 & 0.007  \\
C1D-LSTM & 0.007 & 0.006 & 0.007 & 0.005 & 0.006 & 0.007 & 0.006  \\
C1D-ROC & 0.008 & 0.005 & 0.005 & 0.004 & 0.006 & 0.005 & 0.005  \\
\hline
\end{tabular}
}
\end{center}
\label{DJIA}
\end{table}

\begin{table}[ht!]
\caption{The prediction accuracy in HangSeng Index.}
\begin{center}
\resizebox{1.\hsize}{!}{
\begin{tabular}{c|c|c|c|c|c|c|c}
\hline
Year & Year1 & Year2 & Year3 & Year4 & Year5 & Year6 & Average \\
\hline
\multicolumn{8}{c}{Panel A.MAPE} \\
\hline
WSAEs-LSTM & 0.016 & 0.017 & 0.012 & 0.011 & 0.021 & 0.013 & 0.015  \\
C1D-LSTM & 0.017 & 0.012 & 0.009 & 0.010 & 0.022 & 0.012 & 0.014  \\
C1D-ROC & 0.011 & 0.011 & 0.008 & 0.009 & 0.010 & 0.011 & 0.010  \\
\hline
\multicolumn{8}{c}{Panel B.Correlation coefficient} \\
\hline
WSAEs-LSTM & 0.944 & 0.924 & 0.920 & 0.927 & 0.904 & 0.968 & 0.931  \\
C1D-LSTM & 0.948 & 0.956 & 0.955 & 0.951 & 0.962 & 0.975 & 0.958 \\
C1D-ROC & 0.979 & 0.964 & 0.955 & 0.952 & 0.985 & 0.979 & 0.969  \\
\hline
\multicolumn{8}{c}{Panel C.Theil U} \\
\hline
WSAEs-LSTM & 0.011 & 0.010 & 0.008 & 0.007 & 0.018 & 0.008 & 0.011  \\
C1D-LSTM & 0.012 & 0.008 & 0.006 & 0.007 & 0.015 & 0.008 & 0.009 \\
C1D-ROC & 0.007 & 0.007 & 0.006 & 0.006 & 0.007 & 0.007 & 0.007  \\
\hline
\end{tabular}
}
\end{center}
\label{HangSeng}
\end{table}

\begin{table}[ht!]
\caption{The prediction accuracy in Nifty 50 index.}
\begin{center}
\resizebox{1.\hsize}{!}{
\begin{tabular}{c|c|c|c|c|c|c|c}
\hline
Year & Year1 & Year2 & Year3 & Year4 & Year5 & Year6 & Average \\
\hline
\multicolumn{8}{c}{Panel A.MAPE} \\
\hline
WSAEs-LSTM & 0.020 & 0.016 & 0.017 & 0.014 & 0.016 & 0.018 & 0.017  \\
C1D-LSTM & 0.014 & 0.014 & 0.022 & 0.015 & 0.019 & 0.014 & 0.016  \\
C1D-ROC & 0.012 & 0.009 & 0.010 & 0.008 & 0.008 & 0.007 & 0.009  \\
\hline
\multicolumn{8}{c}{Panel B.Correlation coefficient} \\
\hline
WSAEs-LSTM & 0.895 & 0.927 & 0.992 & 0.885 & 0.974 & 0.951 & 0.937  \\
C1D-LSTM & 0.946 & 0.962 & 0.992 & 0.866 & 0.971 & 0.969 & 0.951  \\
C1D-ROC & 0.973 & 0.968 & 0.903 & 0.996 & 0.960 & 0.988 & 0.964  \\
\hline
\multicolumn{8}{c}{Panel C.Theil U} \\
\hline
WSAEs-LSTM & 0.013 & 0.010 & 0.010 & 0.009 & 0.010 & 0.011 & 0.011  \\
C1D-LSTM & 0.010 & 0.009 & 0.014 & 0.010 & 0.012 & 0.009 & 0.011  \\
C1D-ROC & 0.007 & 0.006 & 0.007 & 0.005 & 0.005 & 0.005 & 0.006  \\
\hline
\end{tabular}
}
\end{center}
\label{Nifty_50}
\end{table}

\begin{table}[ht!]
\caption{The prediction accuracy in Nikkei 225 index.}
\begin{center}
\resizebox{1.\hsize}{!}{
\begin{tabular}{c|c|c|c|c|c|c|c}
\hline
Year & Year1 & Year2 & Year3 & Year4 & Year5 & Year6 & Average \\
\hline
\multicolumn{8}{c}{Panel A.MAPE} \\
\hline
WSAEs-LSTM & 0.024 & 0.019 & 0.019 & 0.019 & 0.018 & 0.017 & 0.019  \\
C1D-LSTM & 0.016 & 0.011 & 0.010 & 0.019 & 0.012 & 0.010 & 0.013  \\
C1D-ROC & 0.013 & 0.010 & 0.013 & 0.010 & 0.013 & 0.013 & 0.012 \\
\hline
\multicolumn{8}{c}{Panel B.Correlation coefficient} \\
\hline
WSAEs-LSTM & 0.878 & 0.834 & 0.665 & 0.972 & 0.774 & 0.924 & 0.841  \\
C1D-LSTM & 0.960 & 0.949 & 0.913 & 0.964 & 0.905 & 0.979 & 0.945  \\
C1D-ROC & 0.957 & 0.972 & 0.994 & 0.943 & 0.981 & 0.969 & 0.969  \\
\hline
\multicolumn{8}{c}{Panel C.Theil U} \\
\hline
WSAEs-LSTM & 0.016 & 0.013 & 0.013 & 0.013 & 0.012 & 0.012 & 0.013  \\
C1D-LSTM & 0.010 & 0.007 & 0.007 & 0.017 & 0.008 & 0.006 & 0.009  \\
C1D-ROC & 0.009 & 0.006 & 0.009 & 0.007 & 0.008 & 0.009 & 0.008 \\
\hline
\end{tabular}
}
\end{center}
\label{Nikkei_225}
\end{table}

\begin{table}[ht!]
\caption{The prediction accuracy in S\&P500 Index.}
\begin{center}
\resizebox{1.\hsize}{!}{
\begin{tabular}{c|c|c|c|c|c|c|c}
\hline
Year & Year1 & Year2 & Year3 & Year4 & Year5 & Year6 & Average \\
\hline
\multicolumn{8}{c}{Panel A.MAPE} \\
\hline
WSAEs-LSTM & 0.012 & 0.014 & 0.010 & 0.008 & 0.011 & 0.010 & 0.011  \\
C1D-LSTM & 0.011 & 0.011 & 0.009 & 0.008 & 0.013 & 0.011 & 0.011  \\
C1D-ROC & 0.010 & 0.009 & 0.008 & 0.006 & 0.008 & 0.007 & 0.008  \\
\hline
\multicolumn{8}{c}{Panel B.Correlation coefficient} \\
\hline
WSAEs-LSTM & 0.944 & 0.944 & 0.984 & 0.973 & 0.880 & 0.953 & 0.946  \\
C1D-LSTM & 0.962 & 0.973 & 0.988 & 0.986 & 0.860 & 0.958 & 0.955  \\
C1D-ROC & 0.965 & 0.979 & 0.988 & 0.982 & 0.949 & 0.976 & 0.973  \\
\hline
\multicolumn{8}{c}{Panel C.Theil U} \\
\hline
WSAEs-LSTM & 0.009 & 0.010 & 0.006 & 0.005 & 0.008 & 0.006 & 0.007  \\
C1D-LSTM & 0.007 & 0.007 & 0.006 & 0.005 & 0.008 & 0.007 & 0.007  \\
C1D-ROC & 0.007 & 0.006 & 0.005 & 0.004 & 0.005 & 0.005 & 0.005  \\
\hline
\end{tabular}
}
\end{center}
\label{SP500}
\end{table}

If past prices are used to predict future stock prices, predicting the rate of change of the price is also able to get the current prices. For most stock price series, the price scale is much larger than the rate of change. If the prediction target for the model is the absolute price, it is easy to ignore the information for price changes because changes in the price has a smaller effect on the loss than the absolute price. Tables \ref{CSI_300}-\ref{SP500} show that the model predicts prices indirectly through predicting the rate of change can get higher accuracy. This demonstrates that predicting the rate of change is a better way than to predict prices directly.

\subsection{Predictive curve}
\begin{figure*}
\begin{center}
\includegraphics[width=182.36mm]{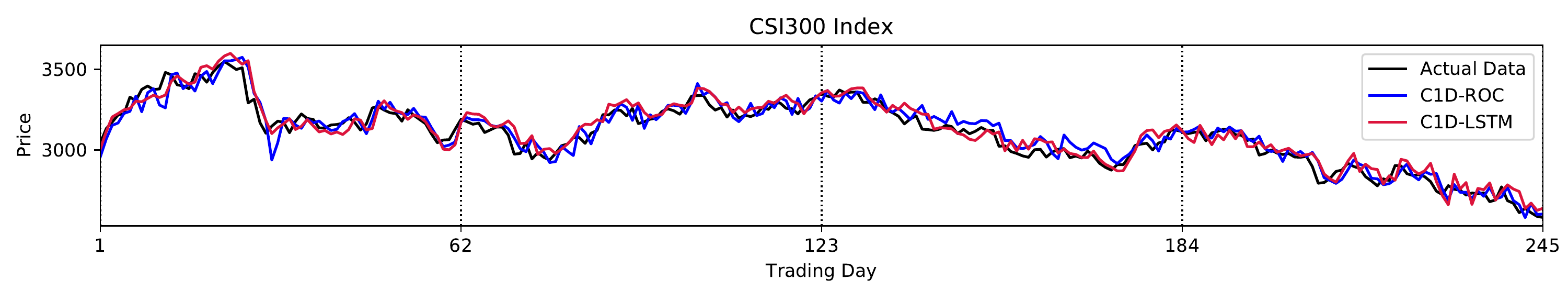}
\includegraphics[width=182.36mm]{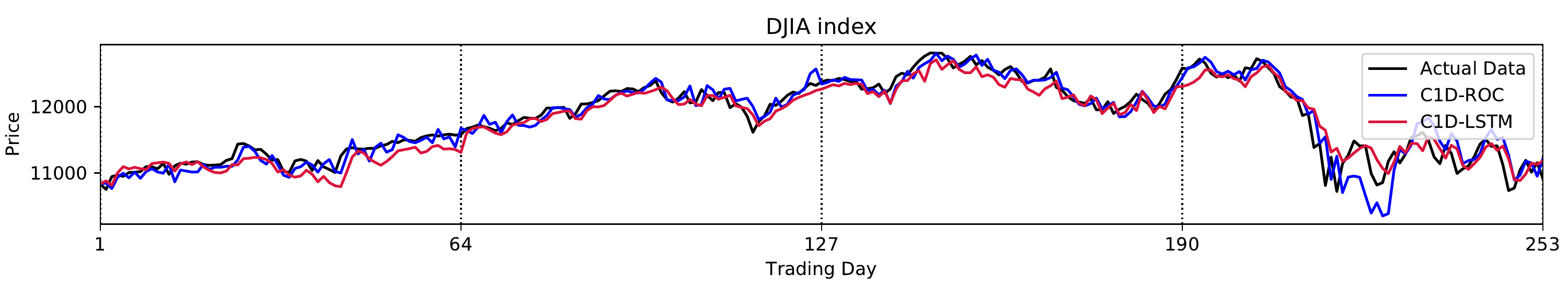}
\includegraphics[width=182.36mm]{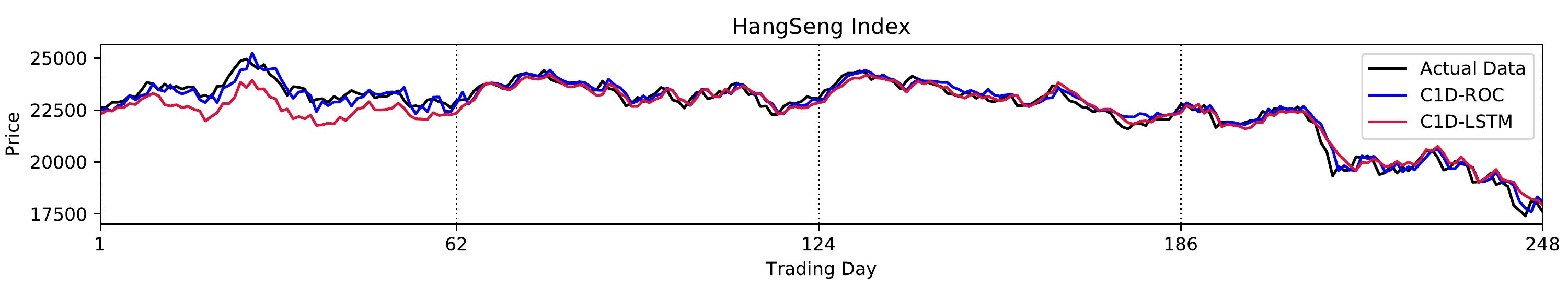}
\includegraphics[width=182.36mm]{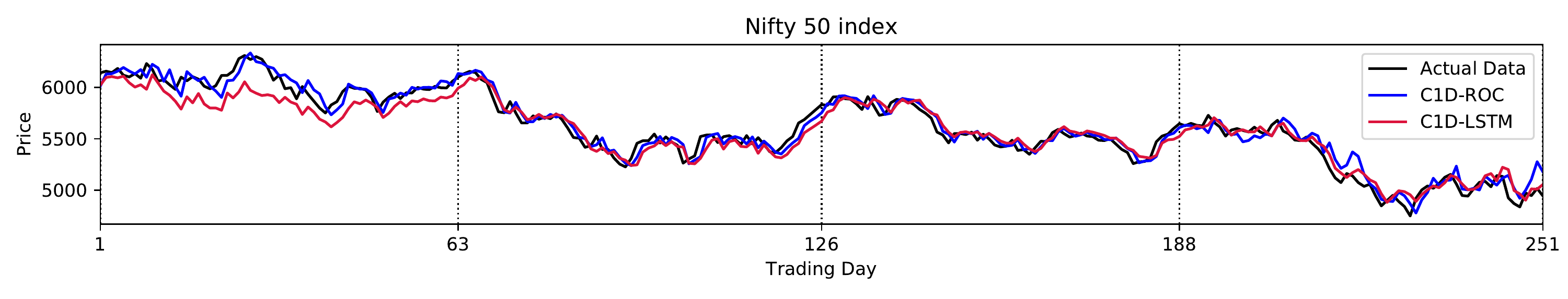}
\includegraphics[width=182.36mm]{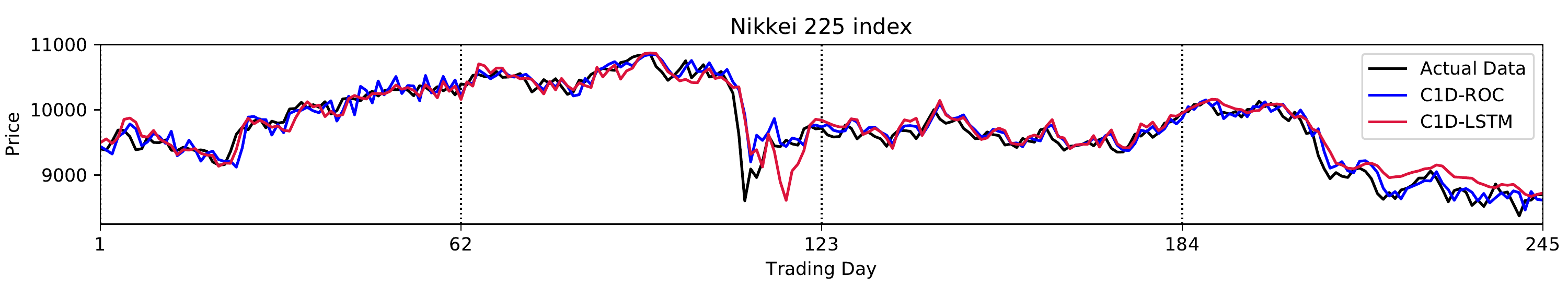}
\includegraphics[width=182.36mm]{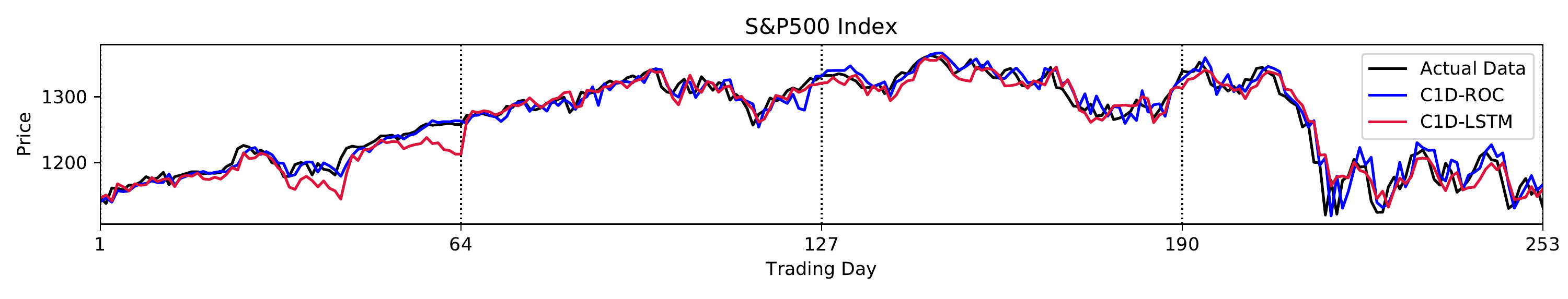}
\end{center}
   \caption{The actual and predicted curves for six stock index from 2010.10.01 to 2011.09.30.}
\label{year1}
\end{figure*}

The predicted results for the first year for each market index are shown in Figure \ref{year1}. The curve for C1D-ROC is closer to the actual curve than that for C1D-LSTM. The curve for C1D-LSTM occasionally deviates far from the actual price curve but that for the C1D-ROC does so only rarely. This demonstrates that future prices can be derived using the current price and price changes. The current input characteristics include the current price but it is difficult to fully preserve this feature in the input features for an autoencoder. If the change in the price is predicted directly and then inferred from the exact current value, the model can use the full information for the current price.

\section{Conclusion}\label{Conclusion}
1-D ResNet sparse autoencoders are used to de-noise and reduce the dimensionality of data. A notional experiment is used to compare the performance of the model that uses features after de-noising and that of a single network with LSTM. The first method reduces over-fitting when there is a lot of noise in the data. The results of experiment show that the proposed method gives a more accurate prediction than WSAEs. This is the first contribution of this paper. Another contribution is that we add prior knowledge about the relationship between prices and the rate of change to the model to try to improve the performance, and the results of experiment show the conclusion that it is more accurate to use the rate of change to indirectly predict the price of stocks than to directly predict the price of stocks.

Future study will use an attention model \cite{chen2017enhancing} to improve the performance. This model assumes that the price for the next day is approximately related to the price for previous days. The attention model will be applied to express the relationship between the price for previous day and next day, which will give improved performance and result that are more easily interpreted.


%





\ifCLASSOPTIONcaptionsoff
  \newpage
\fi



%

\bibliographystyle{IEEEtran}
\bibliography{paper}

%








\begin{figure*}
\begin{center}
\includegraphics[width=182.36mm]{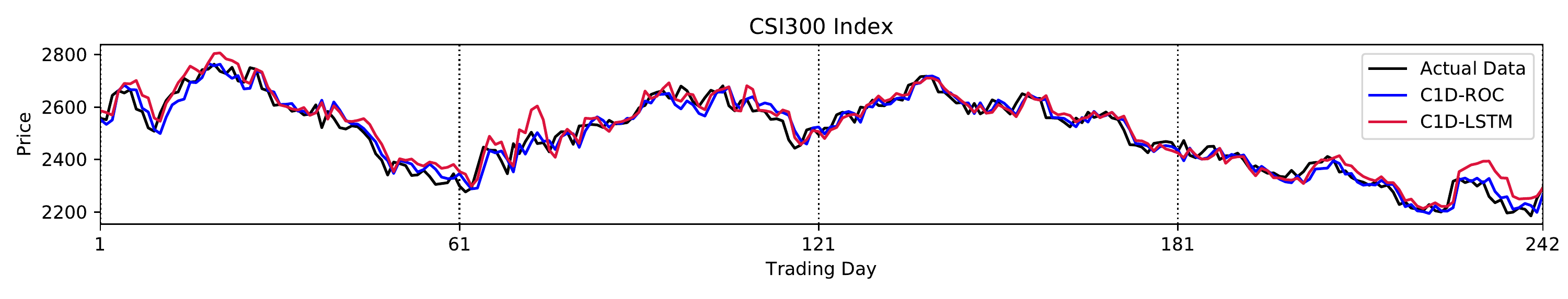}
\includegraphics[width=182.36mm]{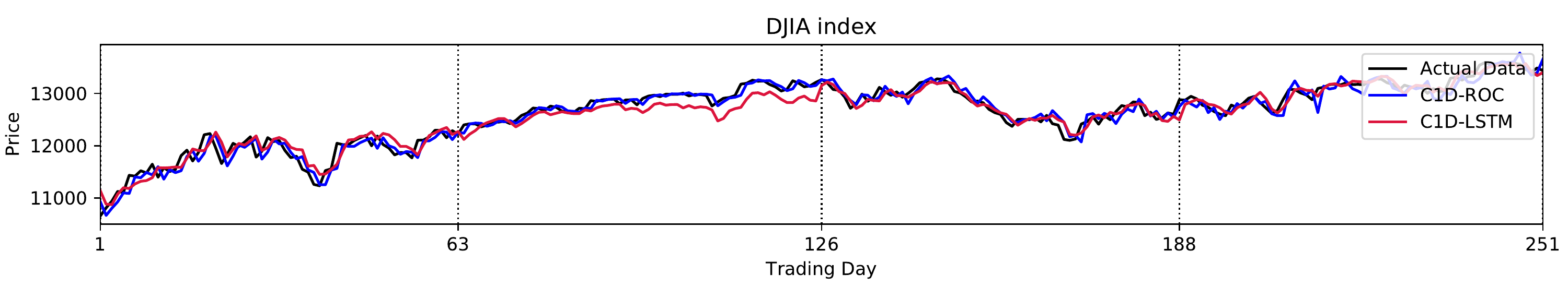}
\includegraphics[width=182.36mm]{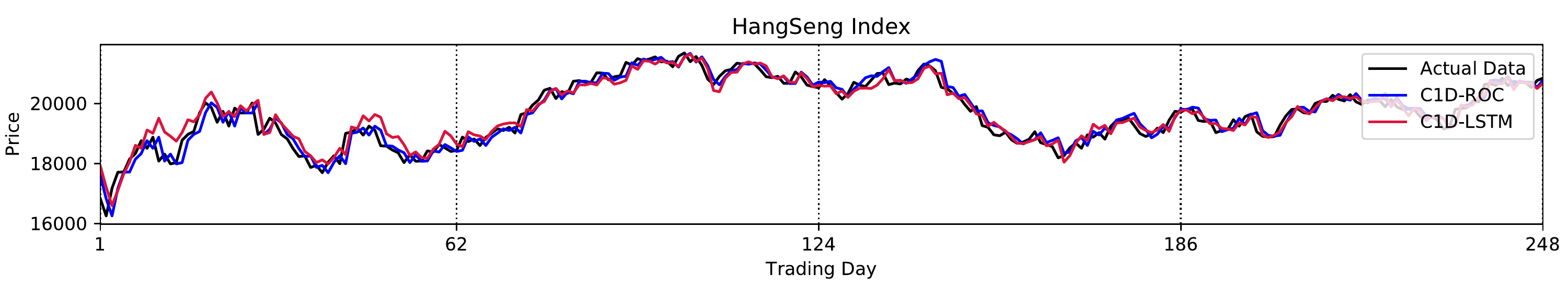}
\includegraphics[width=182.36mm]{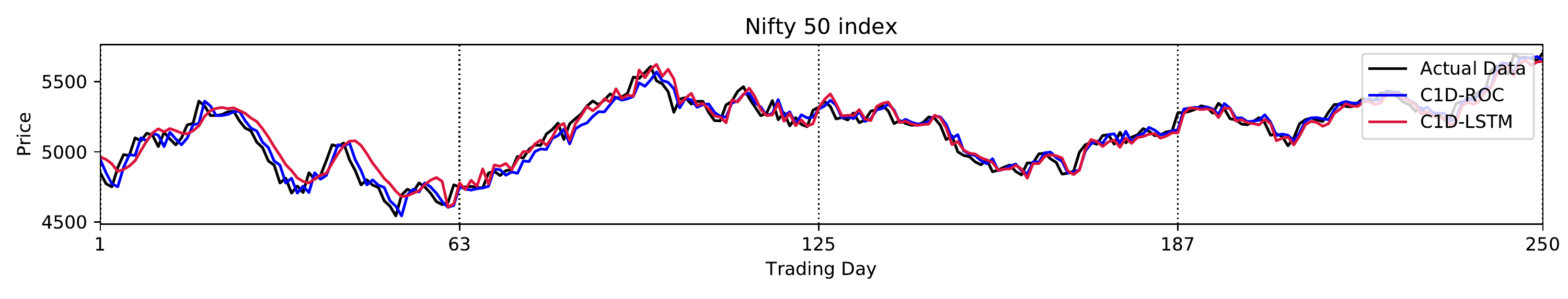}
\includegraphics[width=182.36mm]{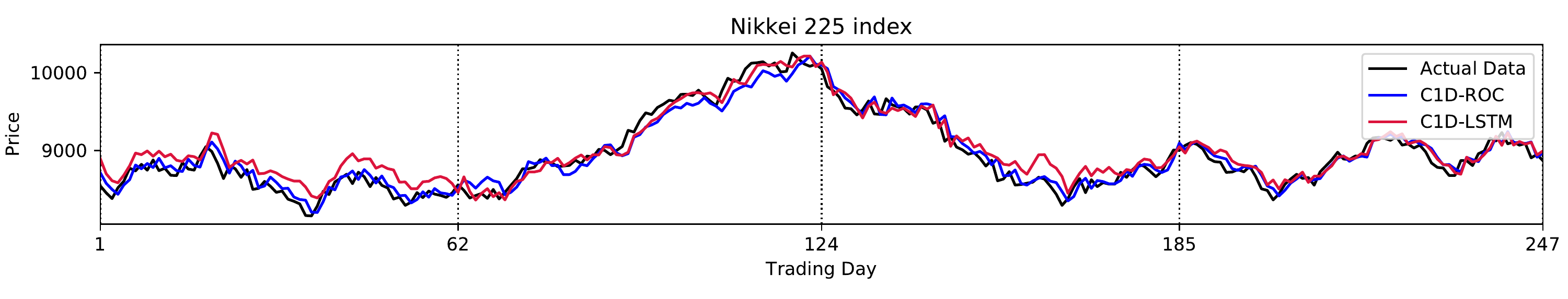}
\includegraphics[width=182.36mm]{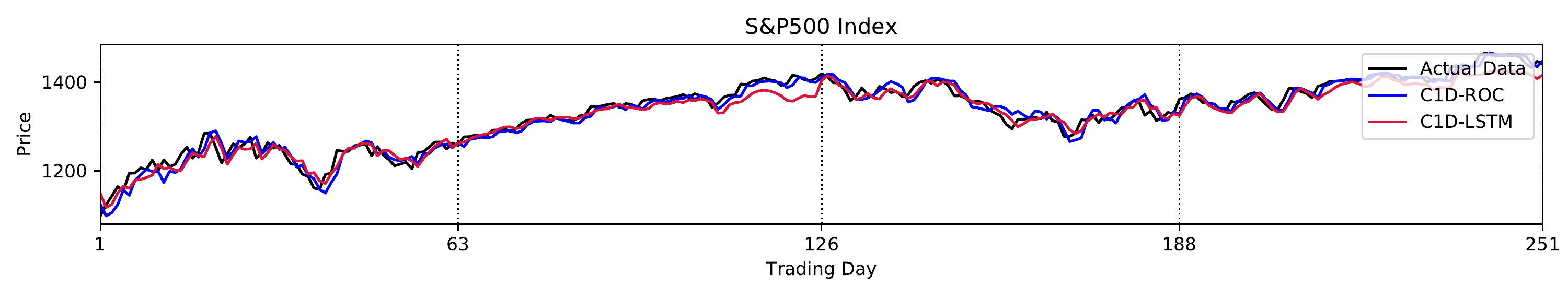}
\end{center}
\caption{The actual and predicted curves for six stock index from 2011.10.01 to 2012.09.30.}
\label{year2}
\end{figure*}

\begin{figure*}
\begin{center}
\includegraphics[width=182.36mm]{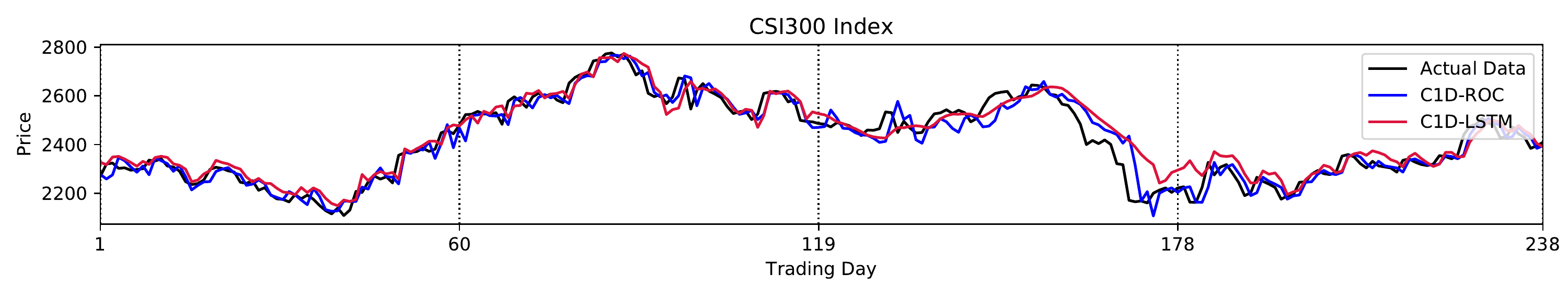}
\includegraphics[width=182.36mm]{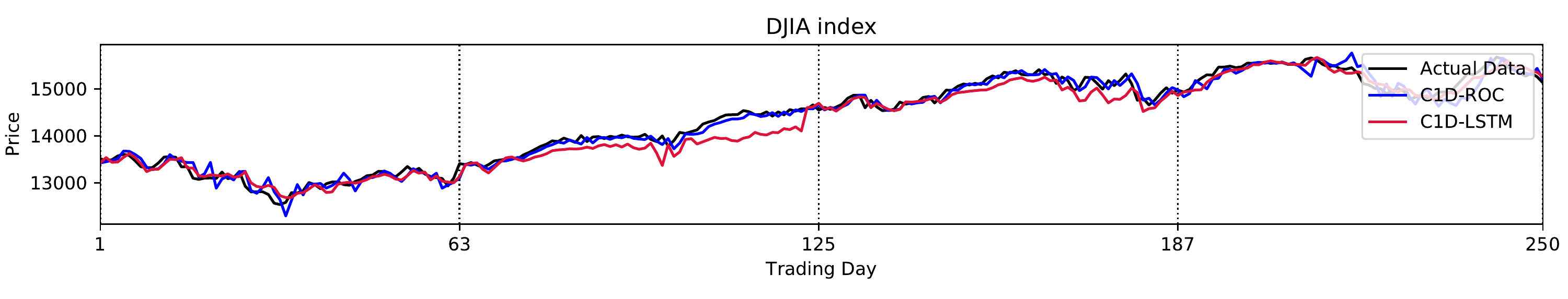}
\includegraphics[width=182.36mm]{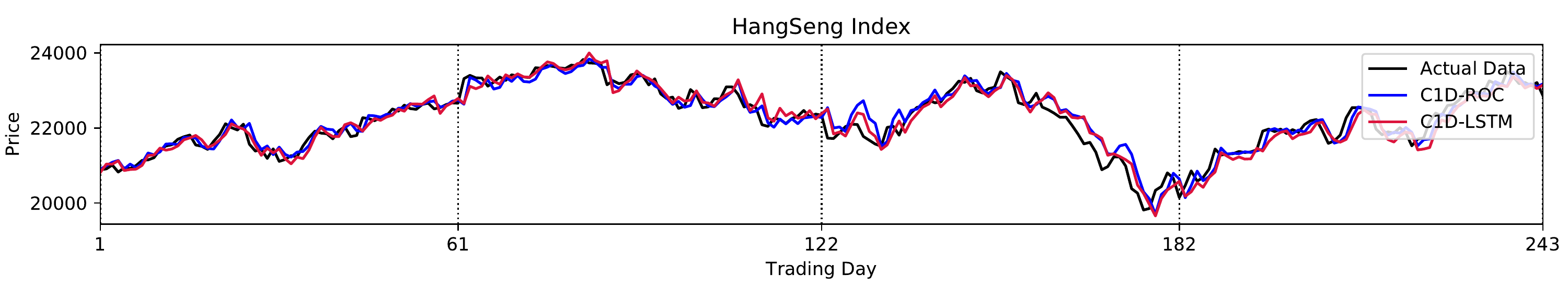}
\includegraphics[width=182.36mm]{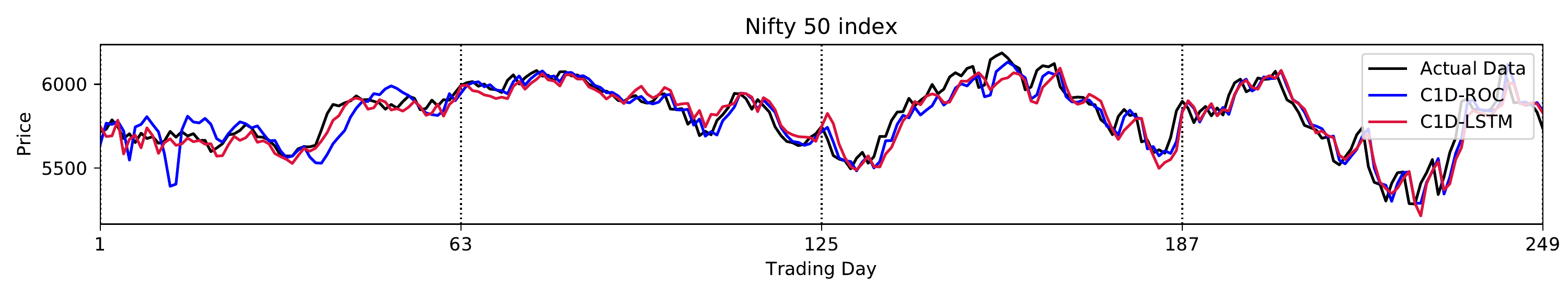}
\includegraphics[width=182.36mm]{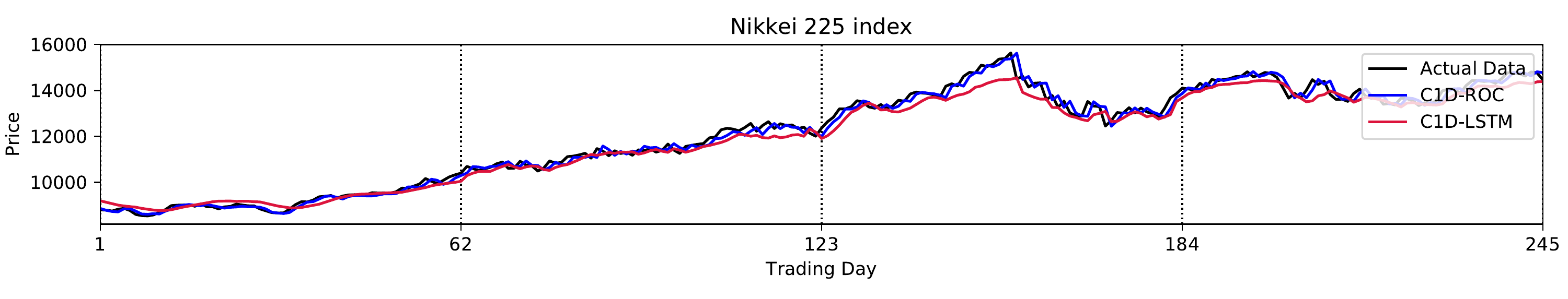}
\includegraphics[width=182.36mm]{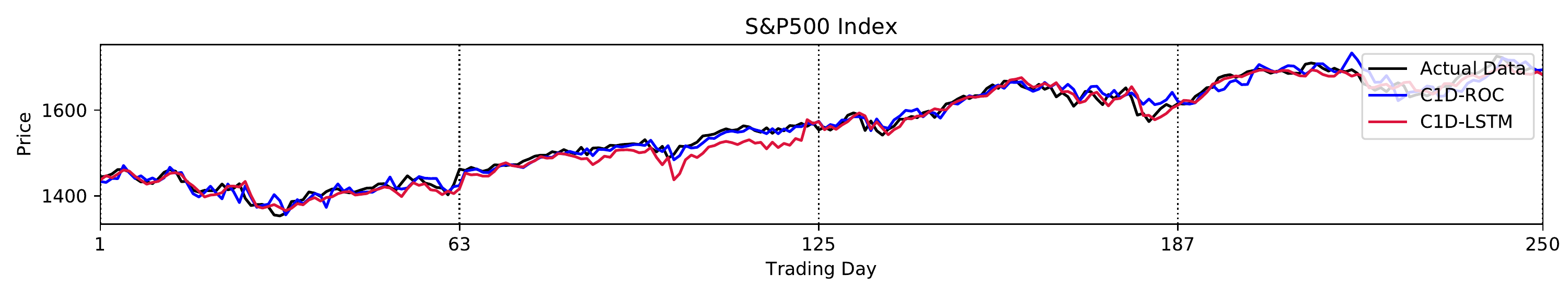}
\end{center}
\caption{The actual and predicted curves for six stock index from 2012.10.01 to 2013.09.30.}
\label{year3}
\end{figure*}

\begin{figure*}
\begin{center}
\includegraphics[width=182.36mm]{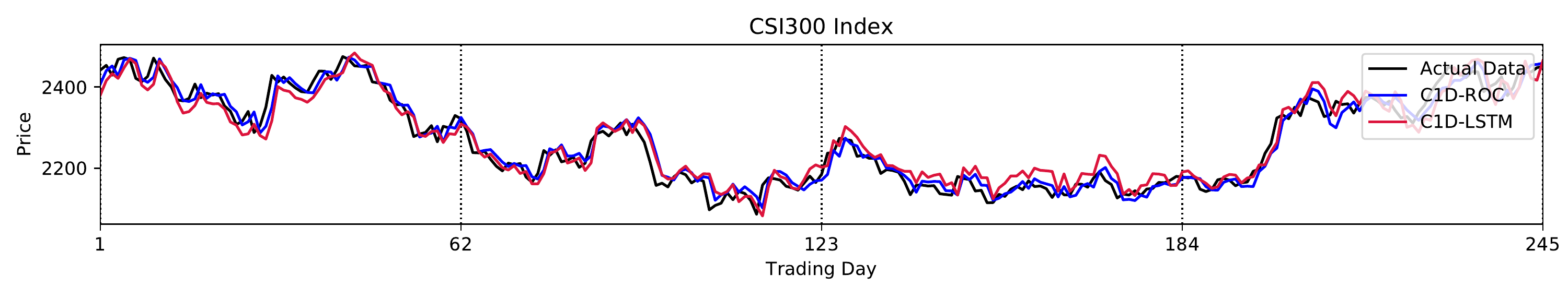}
\includegraphics[width=182.36mm]{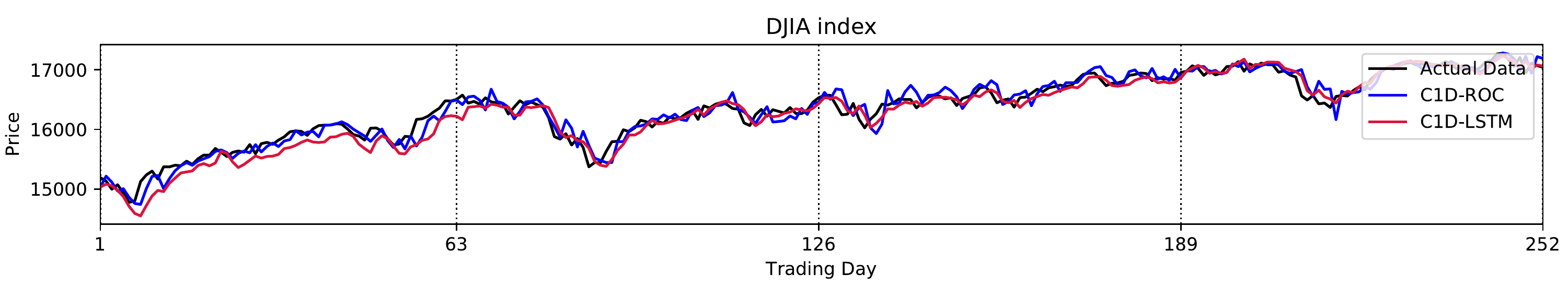}
\includegraphics[width=182.36mm]{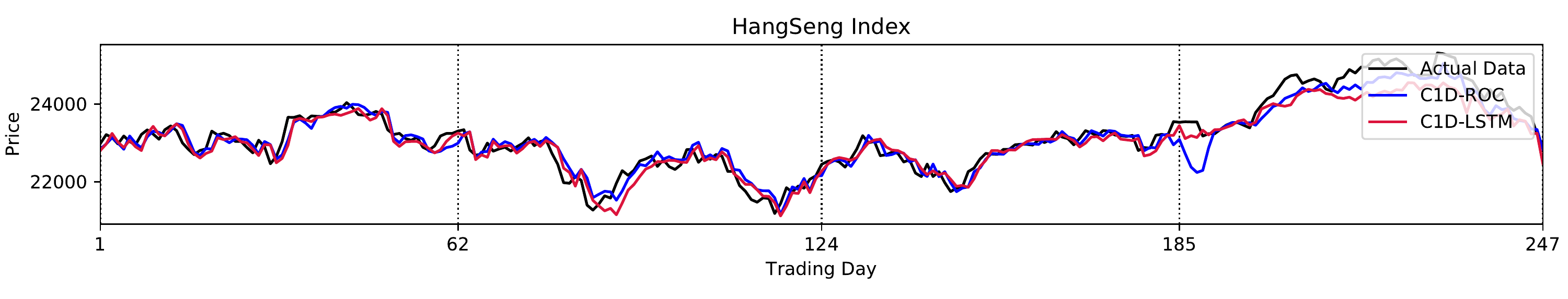}
\includegraphics[width=182.36mm]{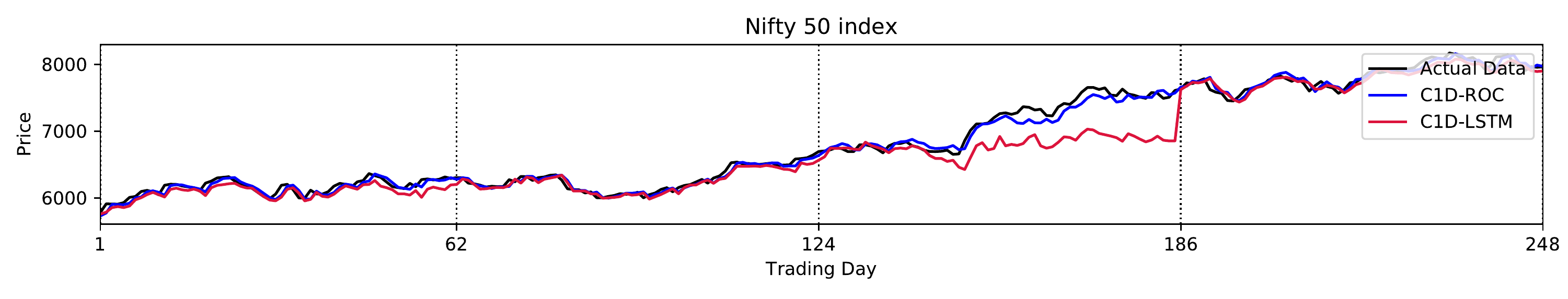}
\includegraphics[width=182.36mm]{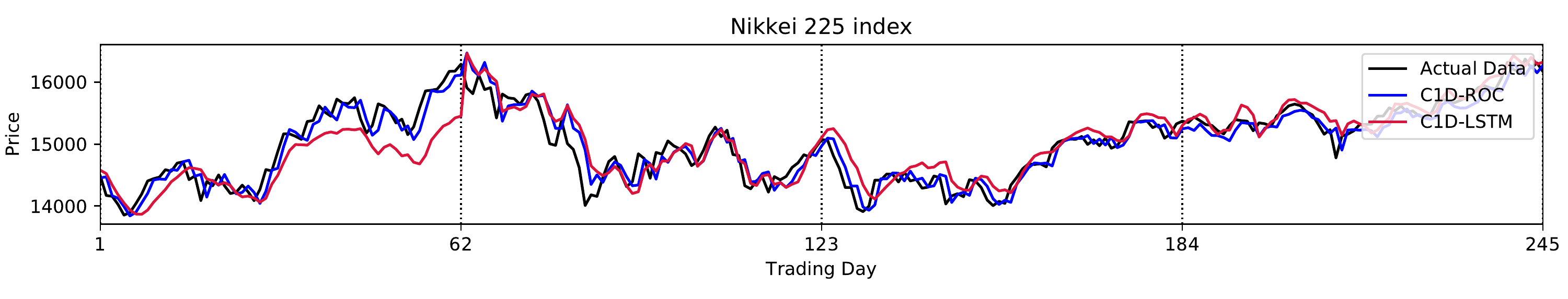}
\includegraphics[width=182.36mm]{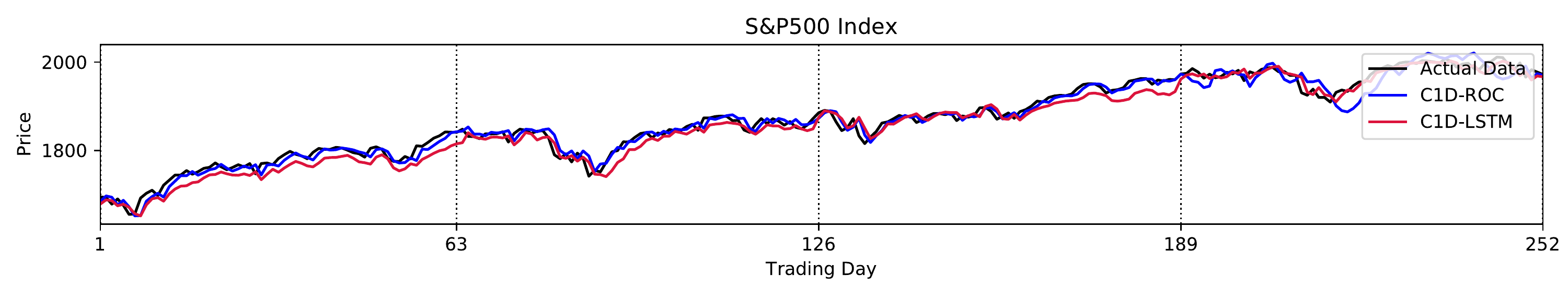}
\end{center}
\caption{The actual and predicted curves for six stock index from 2013.10.01 to 2014.09.30.}
\label{year4}
\end{figure*}

\begin{figure*}
\begin{center}
\includegraphics[width=182.36mm]{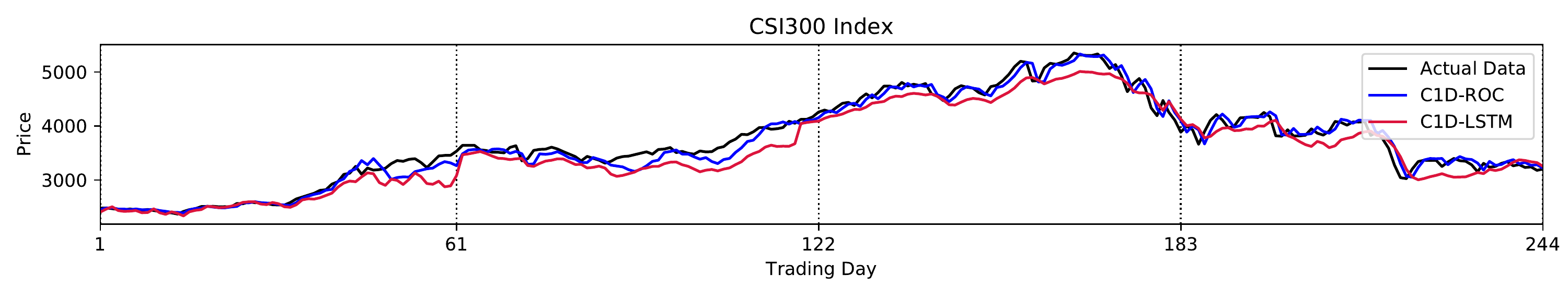}
\includegraphics[width=182.36mm]{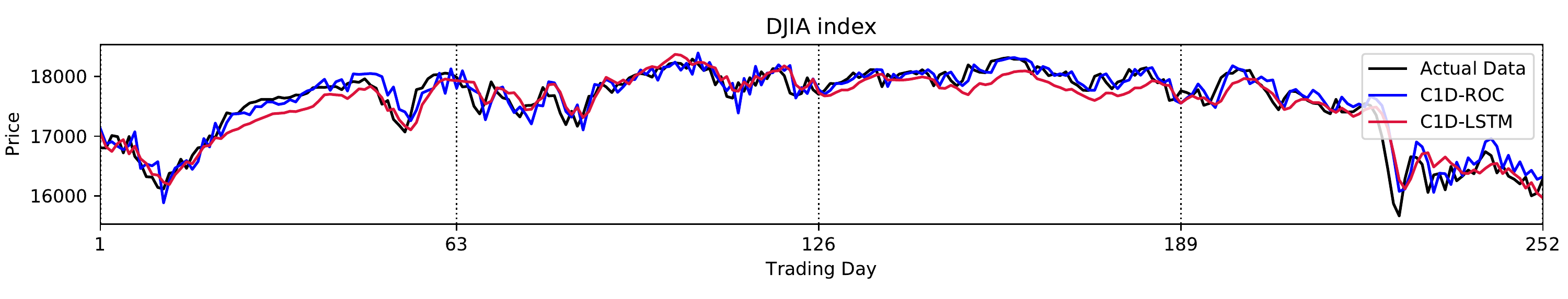}
\includegraphics[width=182.36mm]{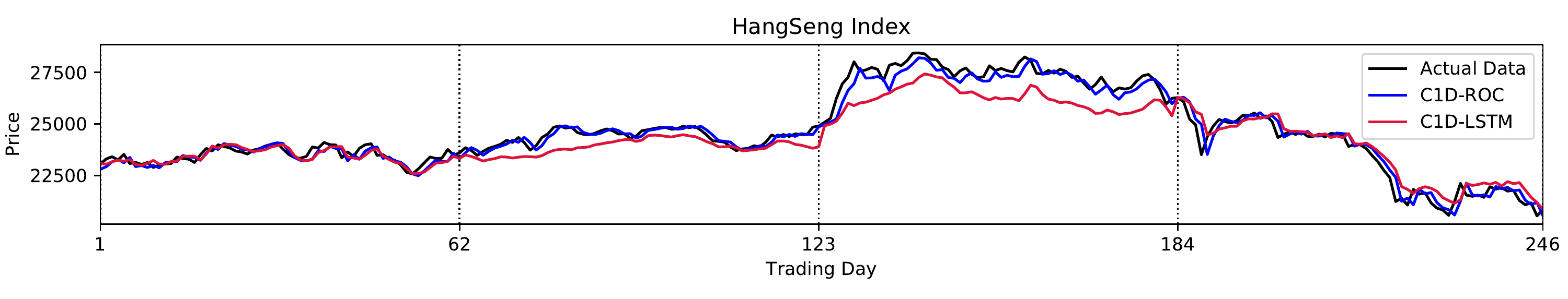}
\includegraphics[width=182.36mm]{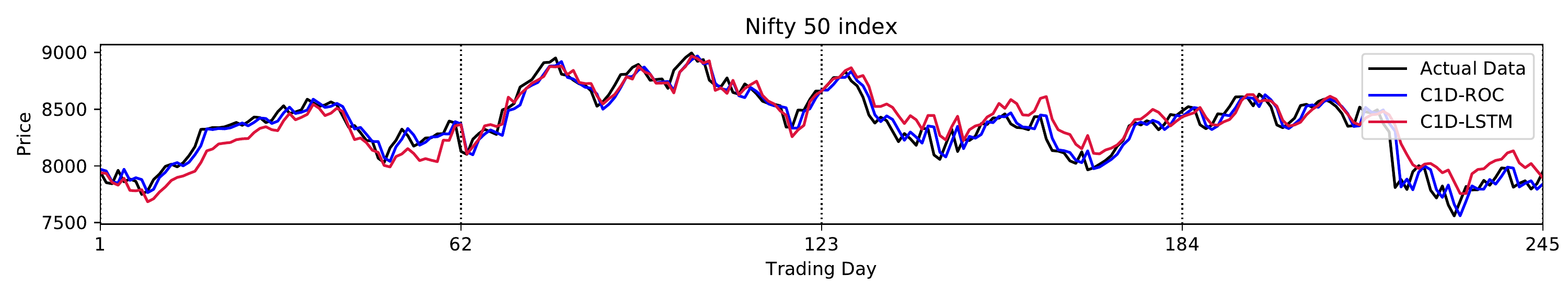}
\includegraphics[width=182.36mm]{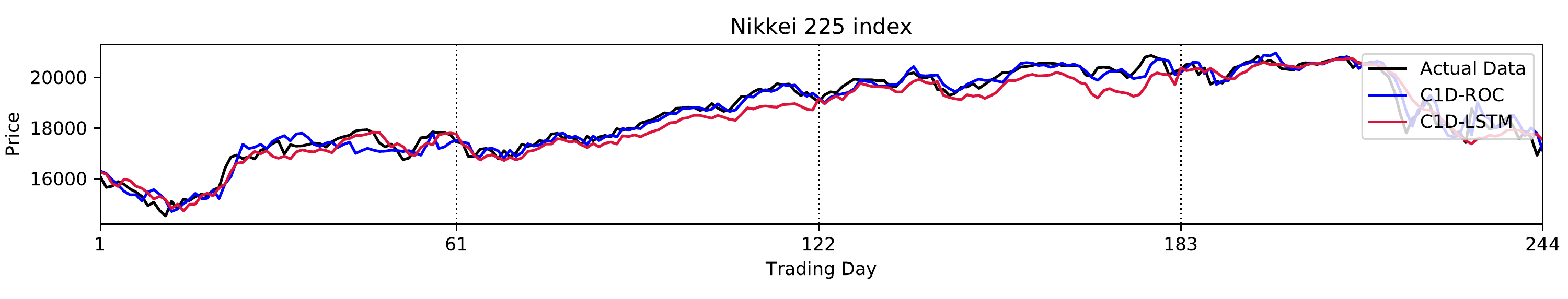}
\includegraphics[width=182.36mm]{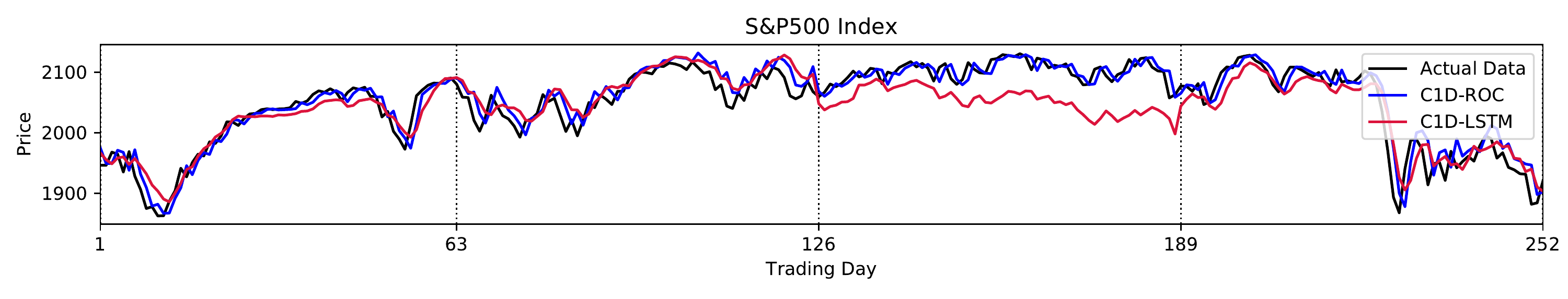}
\end{center}
\caption{The actual and predicted curves for six stock index from 2014.10.01 to 2015.09.30.}
\label{year5}
\end{figure*}

\begin{figure*}
\begin{center}
\includegraphics[width=182.36mm]{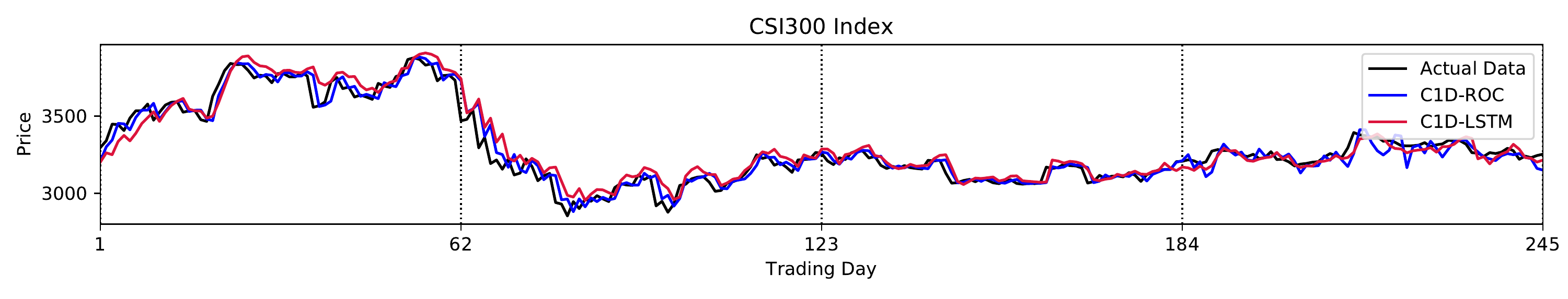}
\includegraphics[width=182.36mm]{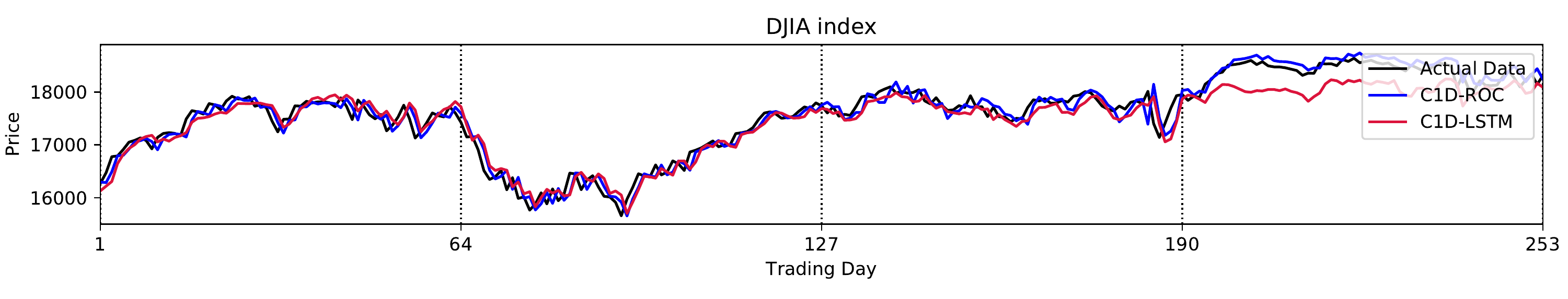}
\includegraphics[width=182.36mm]{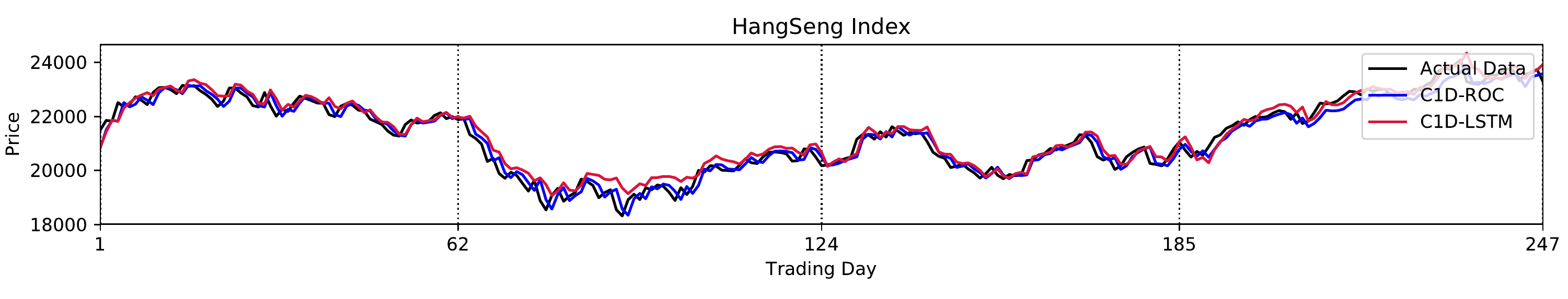}
\includegraphics[width=182.36mm]{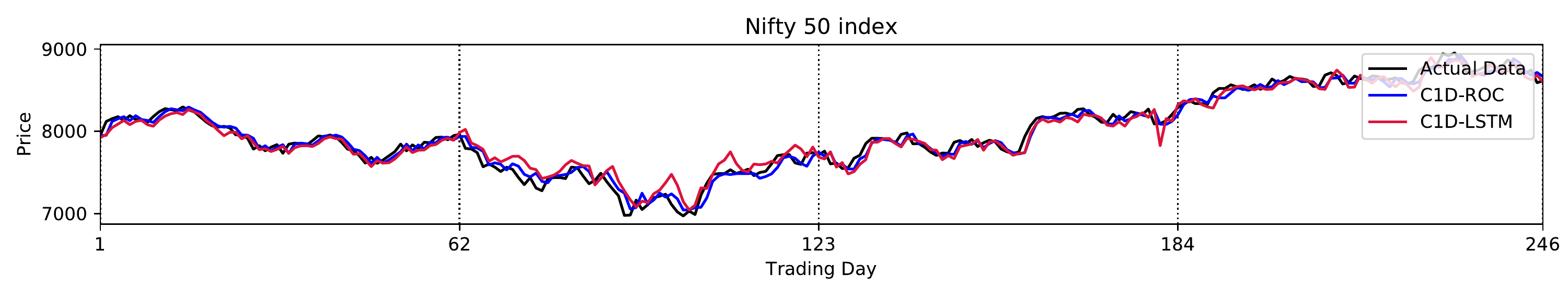}
\includegraphics[width=182.36mm]{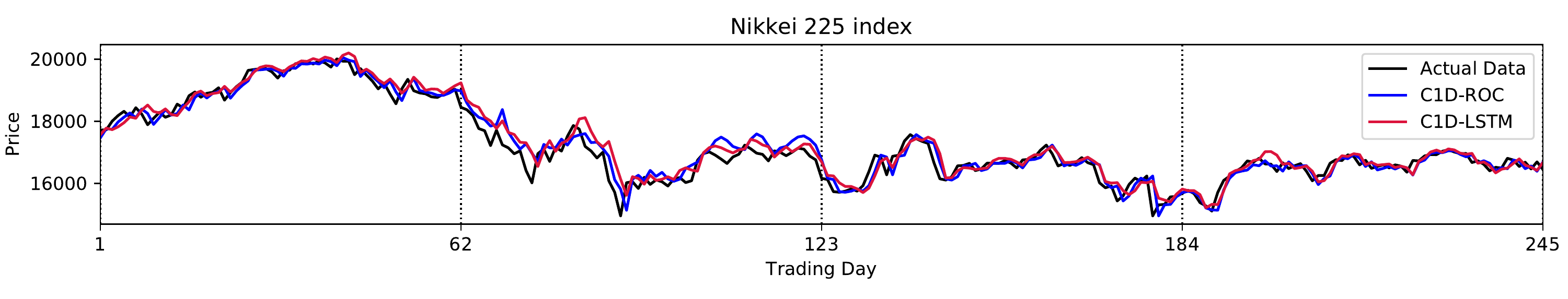}
\includegraphics[width=182.36mm]{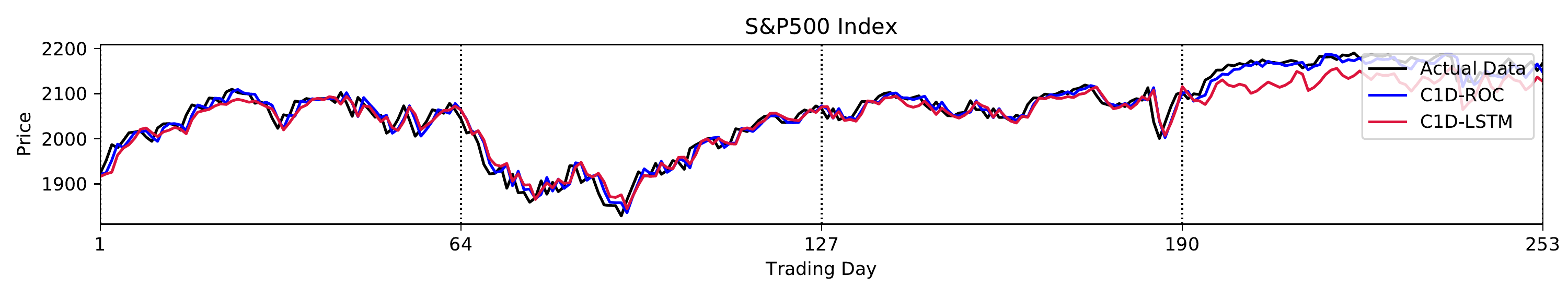}
\end{center}
\caption{The actual and predicted curves for six stock index from 2015.10.01 to 2016.09.30.}
\label{year6}
\end{figure*}

\end{document}